%% file: main-arxiv.tex
\documentclass[letterpaper]{article} 
\usepackage[margin=1in]{geometry}

\usepackage{natbib}
\setcitestyle{square,comma,numbers}

\usepackage{times}

\usepackage[utf8]{inputenc} 
\usepackage[T1]{fontenc}    
\usepackage{hyperref}       
\usepackage{url}            
\usepackage{booktabs}   
\usepackage{multirow}
\usepackage{amsfonts}       
\usepackage{nicefrac}       
\usepackage{microtype}      
\usepackage{xcolor}         

\usepackage{graphicx}
\usepackage{subcaption}
\usepackage{enumitem}
\usepackage{amsmath}
\usepackage{mathtools}
\usepackage[ruled,vlined, linesnumbered]{algorithm2e}
\usepackage{algorithmic}
\usepackage{diagbox}

\newcommand{\A}{A}

\newcommand{\R}{\mathbb{R}}

\newcommand{\E}{\mathbb{E}}
\newcommand{\U}{U}
\newcommand{\LL}{\mathsf{L}}

\newcommand{\cK}{\mathcal{K}}
\newcommand{\cP}{\mathcal{P}}
\newcommand{\cQ}{\mathcal{Q}}
\newcommand{\cW}{\mathcal{W}}

\newcommand{\F}{\mathcal{F}}
\newcommand{\G}{\mathcal{G}}
\newcommand{\Gtrue}{\mathcal{G}^{\dagger}}
\newcommand{\Ldata}{\mathcal{L}_{data}}
\newcommand{\Lpde}{\mathcal{L}_{pde}}
\newcommand{\Lanchor}{\mathcal{L}_{op}}
\newcommand{\Lop}{\mathcal{L}_{op}}

\newcommand{\Jdata}{\mathcal{J}_{data}}
\newcommand{\Jpde}{\mathcal{J}_{pde}}

\newtheorem{definition}{Definition}

\newtheorem{proof*}{Proof}[section]

\title{Physics-Informed Neural Operator\\
for  Learning Partial Differential Equations}

\author{
Zongyi Li*, 
Hongkai Zheng*,
Nikola Kovachki,  
David Jin,
Haoxuan Chen,\\
Burigede Liu, 
Kamyar Azizzadenesheli, 
Anima Anandkumar
}

\date{}

\begin{document}

\maketitle

\input{sections/0-abstract}
\input{sections/1-introduction}

\input{sections/2-problem-setting}
\input{sections/3-method}

\input{sections/5-experiments}

\input{sections/6-discussion}

\section*{Acknowledgement}
The authors want to thank Sifan Wang for meaningful discussions.

Z. Li gratefully acknowledges the financial support from the Kortschak Scholars, PIMCO Fellows, and Amazon AI4Science Fellows programs.
N. Kovachki is partially supported by the Amazon AI4Science Fellowship.
A. Anandkumar is supported in part by Bren endowed chair professorship.

This work was carried out on (1) the NVIDIA NGC as part of Zongyi Li's internship and (2) the Resnick High Performance Computing Center, a facility supported by Resnick Sustainability Institute at the California Institute of Technology

\bibliography{main}
\bibliographystyle{plainnat}

\appendix

\input{sections/Appendix}

\end{document}

%% file: sections/0-abstract.tex
\begin{abstract}
%
In this paper, we propose  physics-informed neural operators (PINO) that combine training data  and physics constraints to learn the  solution operator of a given family of parametric Partial Differential Equations (PDE). PINO is the first hybrid approach incorporating data and PDE constraints at different resolutions to learn the operator. Specifically, in PINO, we combine  coarse-resolution training data with    PDE constraints imposed at a higher resolution. The resulting PINO model can accurately approximate the ground-truth solution operator  for many popular PDE families and shows  no degradation in accuracy even under zero-shot super-resolution, i.e., being able to predict beyond the resolution of training data. 
PINO uses the Fourier neural operator (FNO) framework that is guaranteed to be a universal approximator for any continuous operator and 
discretization convergent in the limit of mesh refinement. By adding PDE constraints to FNO at a higher resolution, we obtain a high-fidelity reconstruction of the ground-truth operator. Moreover, PINO succeeds in   settings where no training data is available and only PDE constraints are imposed,  while previous approaches, such as the  Physics-Informed Neural Network (PINN), fail due to optimization challenges, e.g., in multi-scale dynamic systems such as Kolmogorov flows.

\end{abstract}


%% file: sections/1-introduction.tex
\section{Introduction}

\begin{figure}
    \centering
    \includegraphics[width=0.5\textwidth]{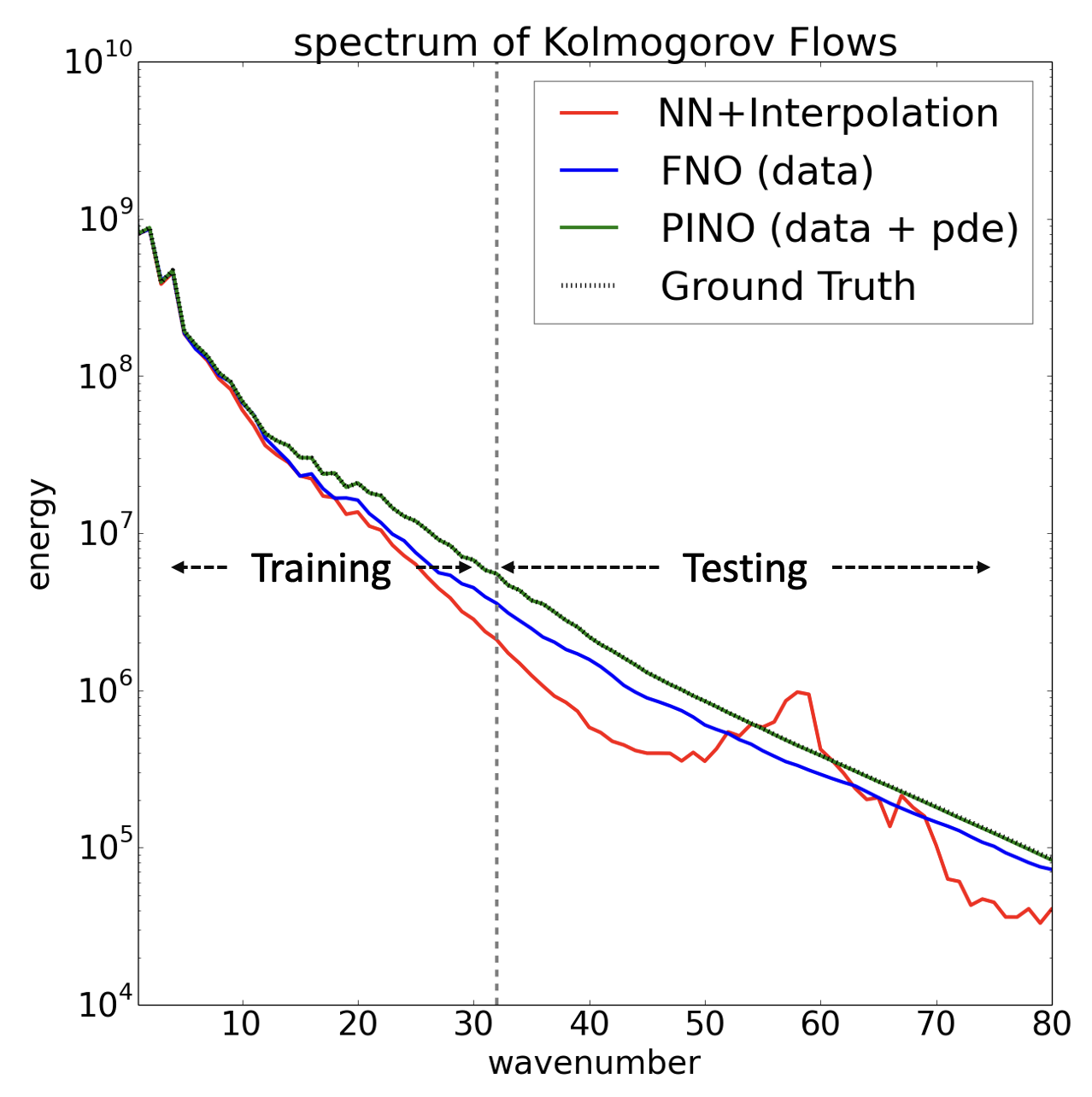}
    \caption{PINO uses both training data and PDE loss function and  perfectly extrapolates to unseen frequencies in Kolmogorov Flows. FNO uses only training data and does not have further information  on higher frequencies, but  still follows the general trend of the ground-truth spectrum. On the other hand, using a trained UNet with trilinear interpolation (NN+Interpolation) has severe distortions at higher frequencies. Details in Section \ref{sec:solve-eqn}.}
    \label{fig:spectrum50}
\end{figure}

Machine learning methods have recently shown promise in solving partial differential equations (PDEs)~\citep{kovachki2021neural, li2020fourier,li2020multipole,lu2019deeponet,  brunton2020machine}.  A recent breakthrough is the paradigm of  operator learning for solving PDEs. Unlike standard neural networks that learn using inputs and outputs of fixed dimensions, neural operators   learn operators, which are mappings between function spaces~\citep{kovachki2021neural, li2020fourier,li2020multipole}. 
The class of neural operators is guaranteed to be  a universal approximator for  any continuous operator~\citep{kovachki2021neural} and hence, has the capacity to approximate any operator including any solution operator of a given family of parametric PDEs. Note that the solution operator is the mapping from the input function (initial   and boundary conditions) to the output solution function. Previous works show that neural operators can capture complex multi-scale dynamic processes and are significantly faster than  numerical solvers~\citep{liu2022learning, yang2022generic, pathak2022fourcastnet, li2022fourier, wen2022accelerating,bonev2023spherical,wen2023real}. 

Neural operators  are proven to be discretization convergent in the limit of mesh refinement~\citep{kovachki2021neural}, meaning they converge to a continuum operator in the limit as the discretization is refined. Consequently, they can be evaluated at any data discretization or resolution at inference time without the need for retraining. For example, neural operators such as Fourier neural operator (FNO) can extrapolate to frequencies that are not seen during training in Kolmogorov Flows, as shown in Figure \ref{fig:spectrum50}, while  standard approaches such as training a UNet and adding trilinear interpolation leads to significantly worse results at higher resolutions.

Even though FNO follows the general trend of the Kolmogorov flow in Figure \ref{fig:spectrum50}, it cannot perfectly match it in the regime of super-resolution, i.e.,  beyond the frequencies seen during training. More generally, neural operators cannot perfectly approximate the ground-truth operator when only coarse-resolution training data is provided. This is a fundamental limitation of  data-driven operator learning methods which depend on  the availability of training data, which can  come either from existing numerical solvers or direct observations of the physical phenomena. In many scenarios, such data can be expensive to generate, unavailable,  or available only as low-resolution observations~\citep{hersbach2020era5}. This limits the ability of neural operators to learn high-fidelity models. Moreover, the generalization of the learned neural operators to unseen scenarios and conditions that are different from training data is challenging.




\subsection{Our Approach and Contributions}

In this paper, we remedy the above shortcomings of data-driven operator learning methods  through the framework of physics-informed neural operators (PINO). Here,  we take a hybrid approach of combining training data (when available) with a PDE loss function at a higher resolution. This allows us to approximate the solution operator of many PDE families nearly perfectly. While there have been many physics-informed approaches proposed previously (discussed in~\ref{sec:related}), ours is the first to incorporate PDE constraints at a higher resolution as a remedy for low resolution training data. We show that this results in high-fidelity solution operator approximations. As shown in Figure \ref{fig:spectrum50}, PINO extrapolates to unseen frequencies in Kolmogorov Flows. Thus, we show that the   PINO model learned from such multi-resolution hybrid  loss functions  has almost no degradation in accuracy even on high-resolution test instances when only low-resolution training data is available. Further, our PINO approach also overcomes the optimization challenges in approaches such as Physics-Informed Neural Network (PINN)~\citep{raissi2019physics} that are purely based on PDE loss functions and do not use training data, and thus, PINO can solve more challenging problems such as time-dependent PDEs.

\begin{figure}
    \centering
    \includegraphics[width=\textwidth]{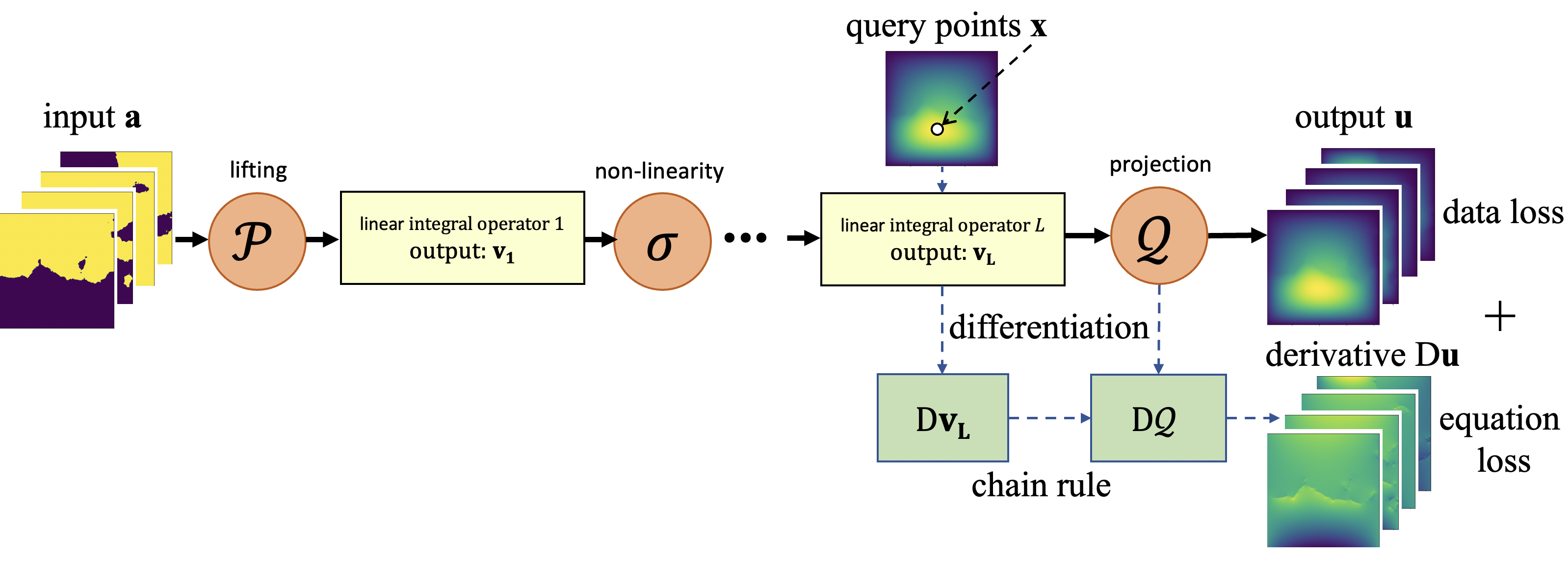}
    \caption{PINO trains neural operator with both training data and PDE loss function. The figure shows the neural operator architecture with the lifting point-wise operator that receives input function $a$ and outputs function $v_0$ with a larger co-dimension. This operation is followed by $L$ blocks that compute linear integral operators followed by non-linearity, and the last layer of which outputs the function $v_L$. The pointwise projection operator projects $v_L$ to output function $u$. Both $v_L$ and $u$ are functions and all their derivatives ($\mathrm{D} v_{\mathrm{L}}$, $\mathrm{D}u$) can be computed in their exact forms at any query points $x$.}
    \label{fig:diagram}
\end{figure}

PINO utilizes both the data and equation loss functions (whichever are available) for operator learning.  To further improve accuracy at test time, we  fine-tune the learned operator on the given PDE instance using only the equation loss function. This allows us to provide a nearly-zero error for the given PDE instance at all resolutions.
A schematic of PINO is shown in Figure \ref{fig:diagram}, where the neural operator architecture is based on Fourier neural operator (FNO) \citep{li2020fourier}. The derivatives needed for the equation loss in PINO are computed explicitly through the operator layers in function spaces. In particular,   we efficiently compute the explicit gradients on function space through Fourier-space computations. In contrast, previous auto-differentiation methods must compute the derivatives at sampling locations. 

The PDE loss function added to PINO vastly improves  \textbf{generalization} and {physical validity} in operator learning compared to purely data-driven methods. PINO requires fewer to no training data and generalizes better compared to the  data-driven FNO~\citep{li2020neural}, especially on high-resolution test instances. On average, the relative error is 7\% lower on both transient and Kolmogorov flows, while matching the speedup of data-trained FNO architecture (400x) compared to the GPU-based pseudo-spectral solver \citep{he2007stability}.  Further, the PINO model on the Navier-Stokes equation can be easily transferred to different Reynolds numbers ranging from $100$ to $500$ using instance-wise fine-tuning.

We also use PINO for solving \textbf{inverse problems} by either: (1) learning the forward solution operator and using gradient-based optimization to obtain the inverse solution, or (2)  learning the inverse solution operator directly. Imposing the PDE loss guarantees the inverse solution is physically valid in both approaches.  We find that of these two approaches, the latter is more accurate for recovering the coefficient function in Darcy flow. We show this approach is 3000x faster than the conventional solvers using accelerated Markov Chain Monte-Carlo (MCMC) \citep{cotter2013mcmc}. 

\subsection{Related Work}\label{sec:related}

Learning solution to PDEs has been proposed under two paradigms: (i) data-driven learning  and (ii) physics-informed optimization. The former  utilizes data from existing solvers or experiments, while the latter is purely based on PDE constraints. An example of data-driven methods  is the class of neural operators  for learning the solution operator of a given  family of parametric PDEs. An example of the physics-based approach is the Physics-Informed Neural Network (PINN) for optimizing the PDE constraints to obtain  the solution function of a given PDE instance. Both these approaches have shortcomings. Neural operators require data, and when that is limited or not available, they are unable to learn the solution operator faithfully. PINN, on the other hand, does not require data but is prone to failure, especially on multi-scale dynamic systems due to optimization challenges.

Neural operators learn the solution operator of a family of PDEs, defined by the map from the input--initial conditions and boundary conditions, to the output--solution functions. In this case, usually, a dataset of input-output pairs from an existing solver or real-world observation is given. There are two main aspects to consider (a) models: the design of models for learning highly complicated PDE solution operators, and (b) data: minimizing data requirements and improving generalization. Recent advances in operator learning replace traditional convolutional neural networks and U-Nets from computer vision with operator-based models tailored to PDEs with greatly improved model expressiveness \citep{li2020neural,lu2019deeponet, patel2021physics,wang2020towards,duvall2021nonlinear}. Specifically, the neural operator generalizes the neural network to the operator setting where the input and output spaces are infinite-dimensional. The framework has successfully approximated solution operators for highly non-linear problems such as turbulence flow \citep{li2020multipole, li2020fourier}. However, the data challenges remain. In particular, (1) training data from an existing solver or an experimental setup is costly to obtain, (2) models struggle in generalizing away from the training distribution, and (3) constructing the most efficient approximation for given data remains elusive. Moreover, it is also evident that in many real-world applications, observational data often is available at only  low resolutions~\citep{hersbach2020era5}, limiting the model's ability to generalize.  



Alternatives to data-driven approaches for solving PDEs  are physics-based approaches that require no training data. A popular framework known as  Physics-Informed Neural Network (PINN)~\citep{raissi2019physics}  uses optimization to find the solution function of a given PDE instance.  PINN uses a neural network as the ansatz of the solution function and optimizes a loss function to minimize the violation of the given equation by taking advantage of auto-differentiation to compute the exact derivatives. PINN overcomes the need to choose a discretization grid that most numerical solvers require, e.g.,  finite difference methods (FDM) and finite element methods (FEM). It has shown promise in solving PDEs for a wide range of applications, including  higher dimensional problems. \citep{lu2021deepxde, han2018solving, hennigh2021nvidia, kashinath2021physics}.
Recently, researchers have developed many variations of PINN with  promising results for solving inverse problems and partially observed tasks \citep{lu202186, zhu2019physics,smith2021hyposvi}. 

However, PINN fails in many  multi-scale dynamic PDE systems~\citep{wang2020and, fuks2020limitations, raissi2020hidden} due to two main reasons, viz., 
 (1) the challenging optimization landscape of the PDE constraints~\citep{wang2021understanding} and its sensitivity to hyper-parameter selection \citep{sun2020surrogate}, and (2) the difficulty in propagating information from the initial or boundary conditions to unseen parts of the interior or to future times~\citep{dwivedi2020physics}. Moreover,  PINN only learns the solution function of a single PDE instance and cannot generalize to other instances without re-optimization. Concurrent work on physics-informed DeepONet that imposes PDE losses on DeepONet~\citep{wang2021learning} overcomes this limitation and can learn across multiple PDE instances. While the PDE loss is computed at any query points, the input is limited to a fixed grid in standard DeepONet~\citep{kovachki2021neural}, and its architecture is a linear method of approximation \citep{lanthaler2022nonlinear}. 
 Our work overcomes all previously mentioned limitations.  Further,  a unique feature that PINO enjoys over other  hybrid learning methods~\citep{zhu2019physics, zhang2021mod, huang2021meta} is its ability  to incorporate data and PDE loss functions at different resolutions. This has not been attempted before, and none of the previous works focus on extrapolation to higher resolutions, beyond what is seen in training data.

%% file: sections/2-problem-setting.tex
\section{Preliminaries and problem settings}
In this section, we first define the stationary and dynamic PDE systems that we consider. We give an overview of the physics-informed setting and operator-learning setting. In the end, we define the Fourier neural operator as a specific model for operator learning.

\begin{figure}
    \centering
    \includegraphics[width=14cm]{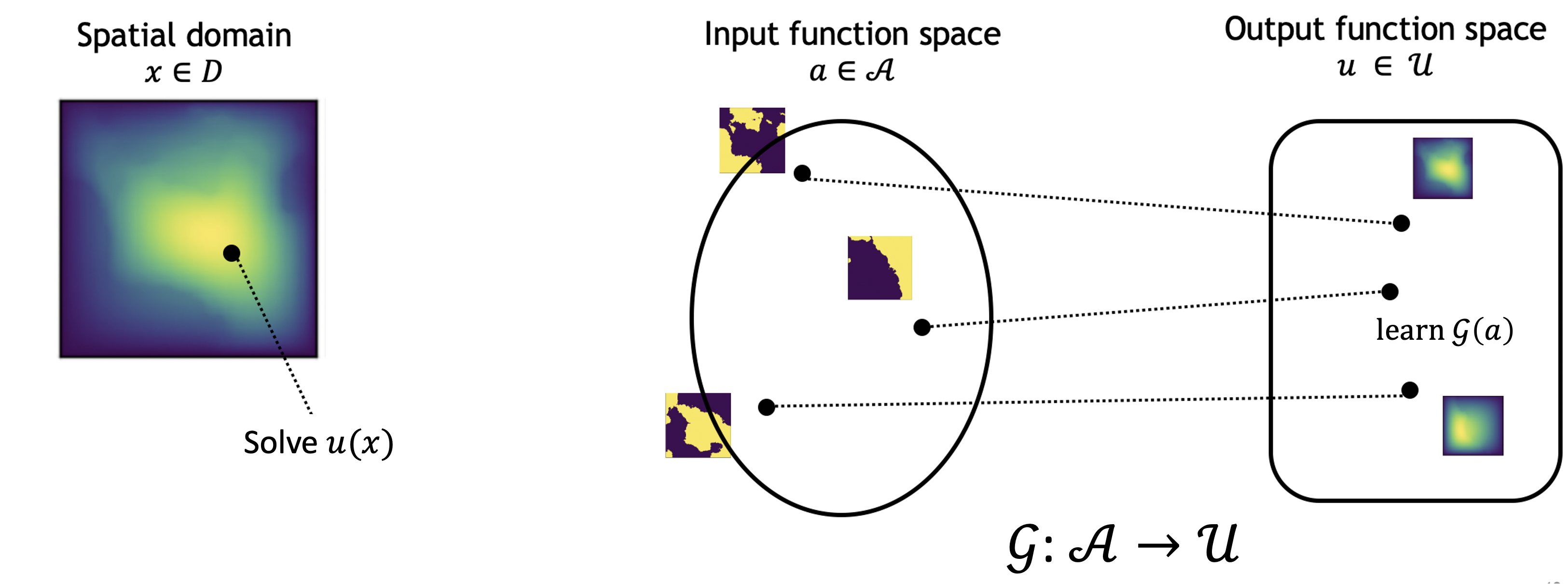}
    \caption{solve for one specific instance verse learn the entire solution operator }
    Left: numerical solvers and PINNs focus on solving one specific instance.
    Right: neural operators learn the solution operator for a family of equations. 
    \label{fig:op}
\end{figure}

\subsection{Problem settings}

We consider two natural classes of PDEs. In the first,
we consider the stationary system
\begin{equation}\label{eq:stationary}
    \begin{split}
    \mathcal{P}(u,a) &= 0, \qquad \text{in } D \subset \R^d \\
    u &= g, \qquad \text{in } \partial D
    \end{split}
\end{equation}
where \(D\) is a bounded domain, \(a \in \mathcal{A} \subseteq \mathcal{V}\) is a PDE coefficient/parameter, \(u \in \mathcal{U}\) is the unknown, and \(\mathcal{P}: \mathcal{U} \times \mathcal{A} \to \mathcal{F}\) is a possibly non-linear partial differential operator with \((\mathcal{U}, \mathcal{V}, \mathcal{F})\) a triplet of Banach spaces. Usually, the function $g$ is a fixed boundary condition but can also potentially enter as a parameter.
This formulation gives rise to the solution operator \(\Gtrue: \mathcal{A} \to \mathcal{U}\) defined by \(a \mapsto u\). A prototypical example is the second-order elliptic equation $ \mathcal{P}(u,a) = -\nabla \cdot (a \nabla u) + f$.


In the second setting, we consider the dynamical system
\begin{align}\label{eq:dynamic}
\begin{split}
    \frac{du}{dt} &= \mathcal{R}(u), \qquad  \text{in } D \times (0,\infty) \\
    u &= g, \qquad \quad \:\:\: \text{in } \partial D \times (0,\infty) \\
    u &= a \qquad \qquad \text{in } \bar{D} \times \{0\}
\end{split}
\end{align}
where \(a = u(0) \in \mathcal{A} \subseteq \mathcal{V}\) is the initial condition, \(u(t) \in \mathcal{U}\) for \(t > 0\) is the unknown, and \(\mathcal{R}\) is a possibly non-linear partial differential operator with \(\mathcal{U}\), and \(\mathcal{V}\) Banach spaces. As before, we take \(g\) to be a known boundary condition. We assume that \(u\) exists and is bounded for all time and for every $u_0 \in \mathcal{U}$.
This formulation gives rise to the solution operator \(\Gtrue : \mathcal{A} \to C \big ( (0,T]; \mathcal{U} \big )\) defined by \(a \mapsto u\). Prototypical examples include the Burgers' equation and the Navier-Stokes equation.

\subsection{Solving equation using the physics-informed neural networks}

Given an instance $a$ and a solution operator \(\Gtrue\) defined by equations \eqref{eq:stationary} or \eqref{eq:dynamic} , we denote by \(u^\dagger = \Gtrue (a)\) the unique ground truth. The equation-solving task is to approximate $u^{\dagger}$. This setting consists of the ML-enhanced conventional solvers such as learned finite element, finite difference, and multigrid solvers \citep{kochkov2021machine,pathak2021ml, greenfeld2019learning}, as well as purely neural network-based solvers such as the Physics-Informed Neural Networks (PINNs), Deep Galerkin Method, and Deep Ritz Method \citep{raissi2019physics,sirignano2018dgm,weinan2018deep}.
Especially, these PINN-type methods use a neural network ${u_{\theta}}$ with parameters ${\theta}$ as the ansatz to approximate the solution function $u^{\dagger}$. The parameters ${\theta}$ are found by minimizing the physics-informed loss with exact derivatives computed using automatic differentiation (autograd). In the stationary case, the physics-informed loss is defined by minimizing the l.h.s. of equation \eqref{eq:stationary} in the squared norm of \(\mathcal{F}\). A typical choice is \(\mathcal{F} = L^2(D)\), giving the loss function
\begin{align}
\label{eq:pinns-stationary}
\begin{split}
    \mathcal{L}_{\text{pde}}(a, u_\theta) &= \Big\|\mathcal{P}(a,u_{\theta})\Big\|^2_{L^2(D)} + \alpha \Big\|u_{\theta}|_{\partial D} - g \Big\|^2_{L^2(\partial D)} \\ 
    &= \int_D |\mathcal{P}(u_{\theta}(x),a(x))|^2\mathrm{d}x  +\alpha\int_{\partial D} |u_{\theta}(x) - g(x)|^2 \mathrm{d}x
\end{split}
\end{align}

In the case of a dynamical system, one minimizes the residual of equation \eqref{eq:dynamic} in some natural norm up to a fixed final time \(T > 0\). A typical choice is the \(L^2 \big( (0,T];L^2(D) \big )\) norm, yielding 
\begin{align}
\label{eq:pinns-dynamic}
\begin{split}
\mathcal{L}_{\text{pde}}(a, u_\theta) 
&= \Big\|\frac{du_{\theta}}{dt} - \mathcal{R}(u_{\theta})\Big\|^2_{L^2(T;D)} + \alpha \Big\|u_{\theta}|_{\partial D} - g \Big\|^2_{L^2(T; \partial D)} + \beta \Big\|u_{\theta}|_{t=0} - a \Big\|^2_{L^2(D)} \\
&= \int_{0}^T \int_D |  \frac{du_{\theta}}{dt}(t,x) -\mathcal{R}(u_{\theta})(t,x)|^2 \mathrm{d}x \mathrm{d}t \\
&\:\:\:+ \alpha \int_{0}^T \int_{\partial D} | u_{\theta}(t,x) - g(t,x)|^2 \mathrm{d}x \mathrm{d}t \\
&\:\:\:+ \beta \int_{D} |u_{\theta}(0,x) - a(x)|^2 \mathrm{d}x
\end{split}
\end{align}
The PDE loss consists of the physics loss in the interior and the data loss on the boundary and initial conditions, with
hyper-parameters \(\alpha, \beta > 0\). Alternatively, the optimization can be formulated via the variational form \citep{weinan2018deep}.

\paragraph{Challenges of PINN}
PINNs take advantage of the universal approximability of neural networks, but, in return, suffer from the low-frequency induced bias. Empirically, PINNs often fail to solve challenging PDEs when the solution exhibits high-frequency or multi-scale structure \citep{wang2021understanding, wang2020and, fuks2020limitations, raissi2020hidden}.
Further, as an iterative solver, PINNs have difficulty propagating information from the initial condition or boundary condition to unseen parts of the interior or to future times \citep{dwivedi2020physics}. 
For example, in challenging problems such as turbulence, PINNs are only able to solve the PDE on a relatively small domain \citep{jin2021nsfnets}, or otherwise, require extra observational data which is not always available in practice \citep{raissi2020hidden, cai2021physics}. In this work, we propose to overcome the challenges posed by optimization by integrating operator learning with PINNs.

\subsection{Learning the solution operator via neural operator}
An alternative setting is to learn the solution operator $\G$.
Given a PDE as defined in \eqref{eq:stationary} or \eqref{eq:dynamic} and the corresponding solution operator \(\Gtrue\), one can use a neural operator ${\G_{\theta}}$ with parameters ${\theta}$ as a surrogate model to approximate \(\Gtrue\). 
Usually we assume a dataset $\{a_j, u_j\}_{j=1}^N$ is available, where $\Gtrue(a_j) = u_j$ and $a_j \sim \mu$ are i.i.d. samples from some distribution \(\mu\) supported on \(\mathcal{A}\). In this case, one can optimize the solution operator by minimizing the empirical data loss on a given data pair
\begin{align}
\label{eq:pinn-data}
\mathcal{L}_{\text{data}}(u, \G_\theta(a)) = \|u - \G_\theta(a)\|_{\mathcal{U}}^2= \int_D |u(x) - \G_\theta(a)(x) |^2 \mathrm{d}x
\end{align}
where we assume the setting of \eqref{eq:stationary} for simplicity of the exposition.
The operator data loss is defined as the average error across all possible inputs
\begin{align}
\label{eq:op-data}
\begin{split}
\mathcal{J}_{\text{data}}( \G_\theta) = \|\Gtrue - \G_\theta\|^2_{L^2_\mu(\mathcal{A};\mathcal{U})} = \E_{a \sim \mu}[ \mathcal{L}_{\text{data}}(a,\theta)] \approx \frac{1}{N} \sum_{j=1}^N \int_D |u_j (x) - {\G_\theta}(a_j)(x) |^2 \mathrm{d}x.
\end{split}
\end{align}
Similarly, one can define the operator PDE loss as
\begin{align}
\label{eq:op-pde}
\begin{split}
\mathcal{J}_{\text{pde}}( \G_\theta) = \E_{a \sim \mu}[ \mathcal{L}_{\text{pde}}(a,  \G_\theta(a)) ].
\end{split}
\end{align}

In general, it is non-trivial to compute the derivatives $d\G_\theta (a)/dx$ and $d\G_\theta (a)/dt$ for model $\G_\theta$. In the following section, we will discuss how to compute these derivatives for Fourier neural operator.

\subsection{Neural operators}
In this work, we will focus on the neural operator model designed for the operator learning problem~\citep{li2020neural}.
The neural operator is formulated as a generalization of standard deep neural networks to the operator setting. Neural operator composes linear integral operator $\cK$ with pointwise non-linear activation function \(\sigma\) to approximate highly non-linear operators.
\begin{definition}[Neural operator $\G_\theta$] Define the neural operator
\begin{equation}
\label{eq:G}
    \G_{\theta} \coloneqq \cQ \circ(\cW_{L} + \cK_{L}) \circ \cdots \circ \sigma(\cW_1 + \cK_1) \circ \cP
\end{equation}
where \(\cP\) and \(\cQ\) are pointwise operators, parameterized with neural networks \(P: \R^{d_a} \to \R^{d_{1}}\) and \(Q: \R^{d_{L}} \to \R^{d_u}\), where ${d_a}$ is the co-dimension of an input function $a\in\A$ and ${d_u}$ is the co-dimension of the output function $u$. \(\cP\) operator lifts the lower dimension function into higher dimensional space and \(\cQ\) operator projects the higher dimension function back to the lower dimensional space. The model stacks $L$ layers of $\sigma(\cW_{l} + \cK_{l})$ where $\cW$ are pointwise linear operators  parameterized as matrices \(W_l \in \R^{d_{{l+1}} \times d_{l}}\) , \(\cK_l: \{D \to \R^{d_{l}}\} \to \{D \to \R^{d_{l+1}}\}\) are integral kernel operators, and \(\sigma\) are fixed activation functions. The parameters $\theta$ consists of all the parameters in $\cP, \cQ, \cW_l, \cK_l$.
\end{definition} 

\begin{definition}[Kernel Integral Operators]
\label{def:kernel}
We define the kernel integral operator \(\cK\) used in \eqref{eq:G}. Let \(\kappa^{(l)} \in C(D \times D; \R^{d_{l+1} \times d_{l}})\) and let \(\nu\) be a Borel measure on \(D\). Then we define \(\cK\) by
\begin{equation}
    \label{eq:kernelop1}
    (\cK v_l)(x) = \int_{D} \kappa^{(l)} (x,y) v_l(y) \: \text{d}\nu(y)
    \qquad \forall x \in D.
\end{equation}
\end{definition}
The kernel integral operator can be discretized and implemented with graph neural networks as in graph neural operators \citep{li2020neural}. 
\begin{equation}
    \label{eq:kernelop2} (\cK v_l)(x) = \sum_{B(x)} \kappa^{(l)}(x,y) v_l(y) \qquad \forall x \in D.
\end{equation}
where $B(x)$ is a ball of center at $x$. As a generalization, the kernel function can also take the form of $(\cK v_l)(x) = \sum_{B(x)} \kappa^{(l)} (x,y, v_l(y)) $.

Recently, \citep{li2020fourier} proposes the Fourier neural operator (FNO) that restricts the integral operator  $\cK$ to convolution. In this case, it can apply the Fast Fourier Transform (FFT) to efficiently compute $\cK$. This leads to a fast architecture that obtains state-of-the-art results for PDE problems.

\begin{definition}[Fourier convolution operator]
\label{def:fourier}
One specific form of the kernel integral operator is the Fourier convolution operator
\begin{equation}
\label{eq:Fourier}
\bigl(\cK v_l\bigr)(x)=   
\F^{-1}\Bigl(R \cdot (\F v_l) \Bigr)(x) \qquad \forall x \in D 
\end{equation}
where $\F, \F^{-1}$ are the Fast Fourier transform and its inverse; $R$ is part of the parameter $\theta$ to be learn. 
\end{definition}
One can build a neural operator with mixed  kernel integral layers and Fourier convolution layers. If the input and output query points are sampled from non-uniform mesh, we can use the graph kernel operator as the first and last layer for continuous evaluation, while using the Fourier layers in the middle for efficient computation, similar to \citep{li2022fourier}.

\paragraph{Challenges of operator learning.}
Operator learning is similar to supervised learning in computer vision and language where data play a very important role. One needs to assume the training points and testing points follow the same problem setting and the same distribution. Especially, the previous FNO model trained on one coefficient (e.g. Reynolds number) or one geometry cannot be easily generalized to another. Moreover, for more challenging PDEs where the solver is very slow or the solver is even not existent, it is hard to gather a representative dataset. 
On the other hand, since prior training methods for FNO do not use any knowledge of the equation, the trained models cannot get arbitrarily close to the ground truth by using the higher resolution as in conventional solvers, leaving a gap of generalization error.
These challenges limit the applications of the prior works beyond accelerating the solver and modeling real-world experiments. In the following section, we will introduce the PINO framework to overcome these problems by using the equation constraints.

\paragraph{Discretization convergent.}
Resolution and discretization convergence is defined as obtaining a unique continuum operator in the limit of mesh refinement \citep{kovachki2021neural}. 
The work \citep{bartolucci2023neural} recently introduced a new concept of representation equivalence, which requires zero aliasing error at each layer, which PINO does not fulfill.  When all the Fourier modes in FNO are active,  an aliasing error is inevitably present. However, in many practical applications, this is typically not an issue, and degraded performance due to aliasing is rarely observed, since the maximum number of modes in an FNO is typically far fewer than the grid size. In fact, the non-linear transformations allow for re-capturing the truncated high-frequency modes which allows for generalization beyond the see training data. Requiring representation equivalence leads to linear methods of approximation which are know to be sub-optimal in their representation capacity \citep{lanthaler2022nonlinear}.

\paragraph{Related work.}
Many machine learning models have been explored for operator learning \citep{Kovachki, nelsen2020random, patel2021physics}.
Besides the above line of work, the deep operator network (DeepONet) is one of the 
 most famous operator models that have shown enormous promise in applications \citep{lu2019deeponet}. 
The work from  \citep{kontolati2023influence} compares the polynomial chaos expansion (PCE), DeepONet, and FNO, and shows that DeepONet has a higher approximation accuracy over PCE. According to Figure 5 in \citep{kontolati2023influence}, standard DeepONet and FNO share a similar convergence rate. A similar comparison study is reported by de Hoop et. al. \citep{de2022cost}  where FNO seems to converge faster. 
We choose FNO as our base model for its scalability to large problems.

%% file: sections/3-method.tex
\section{Physics-informed neural operator (PINO)}
We propose the PINO framework that uses one neural operator model $\G_\theta$ for solving both operator learning problems and equation solving problems. It consists of two phases

\begin{itemize}
    \item \textbf{operator learning}: learn a neural operator $\G_\theta$ to approximate the target solution operator $\Gtrue$ using either/both the data loss $\Jdata$ or/and the PDE loss $\Jpde$. 
    \item \textbf{instance-wise fine-tuning}: use $\G_\theta(a)$ as the ansatz to approximate $u^{\dagger}$ with the pde loss $\Lpde$ and/or an additional operator loss $\Lanchor$ obtained from the operator learning phase.
\end{itemize}

\subsection{Physics-informed operator learning}
For operator learning, we use the physics constraints $\Jpde$ and supervision from data to train the neural operator. Especially one can sample an unlimited amount of virtual PDE instances by drawing additional initial conditions or coefficient conditions $a_j \sim \mu$ for training. In this sense, we have access to the unlimited dataset by sampling new input $a_j$ in each training iteration. This advantage of using PDE constraints removes the requirement on the dataset and makes the supervised problem into a semi-supervised one.

While PINO can be trained with physics constraints $\Jpde$ only, the $\Jdata$ can provide stronger supervision than physics constraints and thus make the optimization much easier. PINO leverages the supervision from any available data to combine with physics constraints for better optimization landscape and thus learning accurate neural operators. A special case is to train a neural operator  on low-resolution data instances with high-resolution PDE constraint, which will be studied in section~\ref{sec:exp}.





\subsection{Instance-wise fine-tuning of trained operator ansatz}
Given a learned operator $\G_\theta$, we use $\G_\theta(a)$ as the ansatz to solve for $u^{\dagger}$. The optimization procedure is similar to PINNs where it computes the PDE loss $\Lpde$ on $a$, except that we propose to use a neural operator instead of a neural network.
Since the PDE loss is a soft constraint and challenging to optimize, we also add an optional operator loss $\Lanchor$ (anchor loss) to bound the further fine-tuned model from the learned operator model 
\[ \Lanchor\big(\G_{\theta_i}(a), \G_{\theta_0}(a)\big) \coloneqq  \|\G_{\theta_i}(a) - \G_{\theta_0}(a)\|_{\mathcal{U}}^2\] 
where $\G_{\theta_i}(a)$ is the model at $i^{th}$ training epoch.
We update the operator $\G_\theta$ using the loss $\Lpde + \alpha \Lanchor$.  It is possible to further apply optimization techniques to fine-tune the last fewer layers of the neural operator and progressive training that gradually increase the grid resolution and use finer resolution in test time.

\paragraph{Optimization landscape.}
Using the operator as the ansatz has two major advantages: (1) PINN does point-wise optimization, while PINO does optimization in the space of functions. In the linear integral operation $\cK$, the operator parameterizes the solution function as a sum of the basis function. Optimization of the set of coefficients and basis is easier than just optimizing a single function as in PINNs. (2) we can learn these basis functions in the operator learning phase which makes the later instance-wise fine-tuning even easier. In PINO, we do not need to propagate the information from the initial condition and boundary condition to the interior. It just requires fine-tuning the solution function parameterized by the solution operator.

\paragraph{Trade-off.} 
(1) complexity and accuracy: 
instance-wise fine-tuning is an option to spend more computation in exchange for better accuracy. The learned operator is extremely fast since it is performing inference on the neural operator. On the other hand, instance-wise fine-tuning can improve accuracy while incurring more computational costs.
(2) resolution effects on optimization landscape and truncation error (i.e. the error of the numerical differentiation): using a higher resolution and finer grid will reduce the truncation error. However, it has a higher computational complexity and memory consumption. A higher resolution may also potentially make the optimization unstable. Using hard constraints such as the anchor loss $\Lanchor$ relieves such a problem.

\subsection{Derivatives of neural operators}
In order to use the equation loss $\Lpde $, one of the major technical challenges is to efficiently compute the derivatives $D(\G_\theta a) =  d(\G_\theta a)/dx$ for neural operators. In this section, we discuss three efficient methods to compute the derivatives of the neural operator  $\G_\theta$ as defined in \eqref{eq:G}.

\paragraph{Numerical differentiation.}
A simple but efficient approach is to use conventional numerical derivatives such as finite difference and Fourier differentiation \citep{zhu2019physics, gao2021phygeonet}. These numerical differentiation methods are fast and memory-efficient: given a $n$-points grid, finite difference requires $O(n)$, and the Fourier method requires $O(n\log n)$. These methods are agnostic to the underlying more. It can be applied to the neural operator with Graph layer \ref{def:kernel} or Fourier layer \ref{def:fourier} or neural networks such as UNet. 

However, the numerical differentiation methods face the same challenges as the corresponding numerical solvers: finite difference methods require a fine-resolution uniform grid; spectral methods require smoothness and uniform grids. Especially. These numerical errors on the derivatives will be amplified on the output solution.

\paragraph{Pointwise differentiation with autograd.} 
Similar to PINN \citep{raissi2019physics}, the most general way to compute the exact derivatives is to use the auto-differentiation library of neural networks (autograd). To apply autograd, one needs to use a neural network to parameterize the solution function $u: x \mapsto u(x)$. However, it is not straightforward to write out the solution function in the neural operator which directly outputs the numerical solution $u = \G_\theta(a)$ on a grid, especially for FNO which uses FFT. To apply autograd, we design a query function $u$ that input $x$ and output $u(x)$.  Recall $\G_{\theta} \coloneqq \cQ \circ  (\cW_{L} + \cK_{L}) \circ \cdots \circ \sigma(\cW_1 + \cK_1) \circ \cP$ and $u = \G_{\theta}a = \cQ v_L = \cQ (\cW_{L} + \cK_{L}) v_{L-1} \ldots$. Since $\cQ$ is pointwise, 
\begin{equation}
\label{eq:autograd}
u(x) = \cQ(v_L)(x) = Q(v_L(x)) = Q \bigl( (\cW_{L} v_{L-1})(x) + \cK_{L}v_{L-1}(x) \bigr)
\end{equation}
For both the kernel operator and Fourier operator, we either remove the pointwise residual term of the last layer $(\cW_{L} v_{L-1})(x)$ or define the query function as an interpolation function on $\cW_{L}$. 

For kernel integral operator \ref{def:kernel}, the kernel function can directly take the query points are input. So the query function 
\[u(x) = Q \bigl( \sum_{B(x)} \kappa^{(l)} (x,y, v_{L-1}(y)) \bigr)\] 
where we omit the derivative of the support $B(x)$. We can apply auto-differentiation to compute the derivatives
\begin{equation}
\label{Du-kernel}
    u'(x) = Q' \bigl(v_L(x)\bigr) \cdot \sum_{B(x)} {\kappa^{(l)}}' (x,y, v_{L-1}(y))
\end{equation}

Similarly, for the Fourier convolution operator, we need to evaluate the Fourier convolution $\cK_{L}v_{L-1}(x)$ on the query points $x$. It can be done by writing out the output function as  Fourier series composing with $Q$ :
\[
u(x) = \cQ \circ 
\F^{-1}\Bigl(R \cdot (\F v_{L-1})) \Bigr)(x) = 
Q \Bigl(
\frac{1}{k_{max}}\sum_{k=0}^{k_{max}} \big(R_k  (\F v_{L-1})_k\big) \exp\frac{i 2\pi k}{D}  (x) 
\Bigr)\]
Where $ \F$ is the discrete Fourier transform. The inverse discrete Fourier transform is the sum of $k_{max}$ Fourier series with the coefficients $\big(R_k  (\F v_{L-1})_k\big)$ coming from the previous layer. 
\begin{equation}
\label{Du-fourier}
    u'(x) = Q'\bigl(v_L(x)\bigr) \cdot \frac{1}{k_{max}}\sum_{k=0}^{k_{max}} \big(R_k  (\F v_{L-1})_k\big) \exp'\frac{i 2\pi k}{D}  (x)
\end{equation}
Notice $\exp'\frac{i 2\pi k}{D}  (x) = \frac{i 2\pi k}{D} \exp\frac{i 2\pi k}{D}  (x)$, just as the numerical Fourier method. If the query points $x$ are a uniform grid, the derivative can be efficiently computed with the Fast Fourier transform.

The autograd method is general and exact, however, it is less efficient. Since the number of parameters $|\theta|$ is usually much greater than the grid size $n$, the numerical methods are indeed significantly faster. Empirically, the autograd method is usually slower and memory-consuming.

\paragraph{Function-wise differentiation.}
While it is expensive to apply the auto differentiation per query point, the derivative can be batched and computed in a function-wise manner.
We develop an efficient and exact derivatives method based on the architecture of the neural operator that can compute the full gradient field. The idea is similar to the autograd, but we explicitly write out the derivatives on the Fourier space and apply the chain rule.
Given the explicit form \eqref{Du-fourier}, $u'$ can be directly computed on the Fourier space. 
\begin{align}
    u' 
    =   \cQ'\bigl(v_L \bigr) \cdot 
    \F^{-1}\Bigl(\frac{i 2\pi }{D}K \cdot (\F v_{L})) \Bigr)
\end{align}
Therefore, to exactly compute the derivative of the Fourier neural operator, one just needs to run the numerical Fourier differentiation. Especially the derivative and be efficiently computed with the Fast Fourier transform when the query is uniform. Similarly, if the kernel function $\kappa^{(l)}$ in \eqref{Du-kernel} has a structured form, we can also write out its gradient field explicitly.

Next, we show how to compute higher orders derivatives in their exact form, without evoking the autograd method. To this end, we can directly apply the chain rule for the higher-order derivatives without calling autograd.  For example, the first order derivatives is $u' = (Q \circ v_L)' = \cQ'(v_L) \cdot v'_L $ and  the 2nd-order derivatives is $u'' = (\cQ v_L)'' =  v'^2_L \cdot \cQ''(v_L)  +  \cQ'(v_L) \cdot v''_L$. Higher-order derivatives can be similarly computed using the chain rule. Furthermore, derivative-based quantities on $v_L$, e.g., $v'_L$ can be computed in its exact form in the Fourier domain. 
Similarly, we can write out the higher-order derivatives of $Q$ using the chain rule. Usually $Q$ is parameterized as a two layer neural networks  $Q(x) = (A_2 \sigma(A_1 x + b_1 ) + b_2)$. So $Q'(x) = A_2 \sigma'(A_1 x + b_1) A_1 $. In this manner,  we have got the explicit formula of the derivatives for all neural operators.

\paragraph{Fourier continuation.}
The Fourier method has its best performance when applied to periodic problems. If the target function is non-periodic or non-smooth, the Fourier differentiation is not accurate. To deal with this issue, we apply the Fourier continuation method that embeds the problem domain into a larger and periodic space.
The extension can be simply done by padding zeros in the input. The loss is computed at the original space during training. The Fourier neural operator will automatically generate a smooth extension.
The details are given in Appendix \ref{app:fc}.


\subsection{Inverse problem}\label{subsec:inverse_problem}
The physics-informed method can be used to solve the inverse problem, where given the output function $u$, the goal is to recover (a distribution of) the input function $a$. By imposing the constraint $\mathcal{P}(u, a)= 0$, we can restrict $a$ to a physically valid manifold. We propose two formulations to do the optimization-based inverse problem with PINO: the forward operator model and the inverse operator model.

\begin{itemize}
\item \textbf{Forward model}: learn the forward operator $\G_\theta: a \mapsto u$ with data. Initialize $\hat{a}$ to approximate $ a^{\dagger}$. Optimize $\hat{a}$ using
\begin{equation}
    \label{eq:forward}
    \mathcal{J}_{forward} \coloneqq \Lpde(\hat{a}, u^\dagger) +\Ldata(\G_\theta(\hat{a})) + R(\hat{a}).
\end{equation}
\item \textbf{inverse model}: learn the inverse operator $\F_\theta: u \mapsto a$ with data. Use $\F_\theta(u^{\dagger})$ to approximate $ a^{\dagger}$. Optimize $\F_\theta$ using 
\begin{equation}
    \label{eq:backward}
    \mathcal{J}_{backward} \coloneqq \Lpde(\F_\theta(u^{\dagger}), u^\dagger) + \Lop(\F_\theta(u^{\dagger}), \F_{\theta_0}(u^{\dagger})) + R(\F_\theta(u^{\dagger})
\end{equation}
\end{itemize}

Where $\Lpde$ is the PDE loss; $\Lanchor$ is the operator loss from the learned operator; $R(a)$ is the regularization term. We use the PDE loss $\Lpde$ to deal with the small error in $\G_\theta$ and the ill-defining issue of $\F_\theta$. We provide a numerical study in section \ref{sec:exp-inverse-problem}.

%% file: sections/5-experiments.tex
\section{Experiments}
\label{sec:exp}

In this section, we conduct empirical experiments to examine the efficacy of the proposed PINO. We present the PDE settings, their domains, and function spaces. In~\ref{sec:operator-learning}, we show using PDE constraint in operator learning results in neural operators that (1) generalize to significantly high-resolution unseen data. 
(2) achieve smaller generalization errors with fewer to no data. Then in~\ref{sec:solve-eqn}, we investigate how PINO uses the operator ansatz to solve harder equations with improved speed and accuracy.  We study three concrete cases of PDEs on Burgers' Equation, Darcy's Equation, and Navier-Stokes equation.  
In~\ref{sec:exp-inverse-problem} we study the inverse problems.
The implementation details of PINO and baseline methods are listed in Appendix \ref{sec:appendix}.


\begin{figure}
    \centering
    \includegraphics[width=0.45\textwidth]{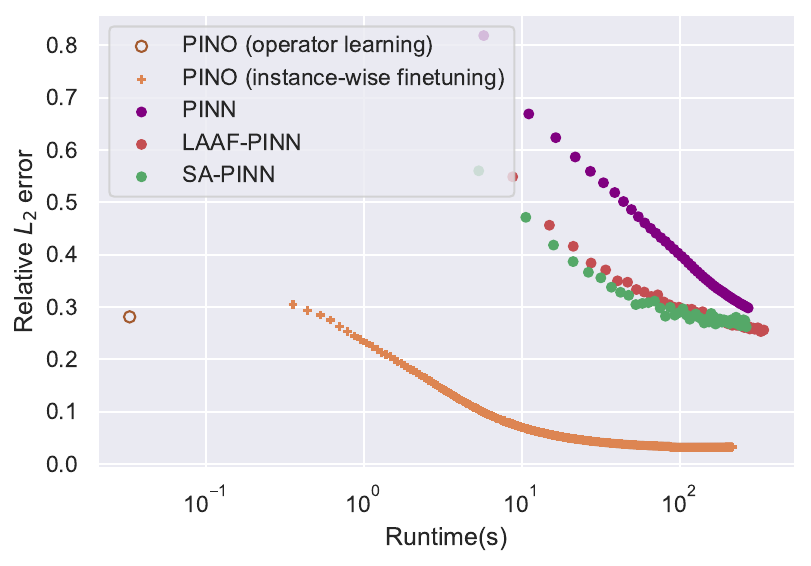}
    \includegraphics[width=0.45\textwidth]{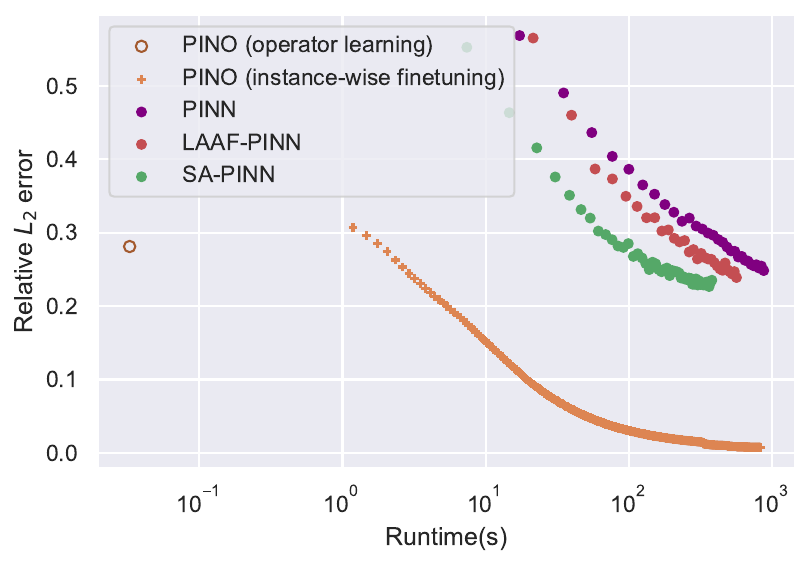}
    \caption{Plot of test relative $L_2$ error versus runtime step for the Kolmogorov flow with Re500, T=0.5s. Left: resolution $64\times 64 \times 65$; right: resolution $128\times 128\times 129$. Averaged over 20 instances. LAAF-PINN: PINN with locally adaptive activation functions. SA-PINN: self-adaptive PINN.}
    \label{fig:pinn-baselines}
\end{figure}

\paragraph{Burgers' Equation.} The 1-d Burgers' equation is a non-linear PDE 
with periodic boundary conditions where $u_0 \in L^2_{\text{per}}((0,1);\R)$ is the initial condition and $\nu = 0.01$ is the viscosity coefficient. We aim to learn the operator mapping the initial condition to the solution, $\Gtrue:u_0 \mapsto u|_{[0,1]}$.
\begin{align}
\label{eq:burgers}
    \begin{split}
    \partial_t u(x,t) + \partial_x ( u^2(x,t)/2) &= \nu \partial_{xx} u(x,t), \qquad x \in (0,1), t \in (0,1] \\
    u(x,0) &= u_0(x), \qquad \qquad \:\: x \in (0,1)
    \end{split}
\end{align}



\paragraph{Darcy Flow.} The 2-d steady-state Darcy Flow equation on the unit box which is the second order linear elliptic PDE 
with a Dirichlet boundary where $a \in L^\infty((0,1)^2;\R_+)$  is a piecewise constant diffusion coefficient and $f = 1$ is a fixed forcing function. We are interested in learning the operator mapping the diffusion coefficient to the solution, 
$\Gtrue: a \mapsto u$. Note that although the PDE is linear, the operator $\Gtrue$ is not.
\begin{align}\label{eq:darcy}
\begin{split}
- \nabla \cdot (a(x) \nabla u(x)) &= f(x) \qquad x \in (0,1)^2 \\
u(x) &= 0 \qquad \quad \:\:x \in \partial (0,1)^2
\end{split}
\end{align}
Since $a$ is in $L_{inf}$, we considered both the strong form $\Lpde(u) = \nabla \cdot (a \nabla u) - f$ and the weak form minimization loss $\Lpde(u) = -\frac{1}{2}(a \nabla u,  \nabla u) - (u, f), u \in H_1$. Experiments show the strong form has a better performance.



\paragraph{Navier-Stokes Equation.}
We consider the 2-d Navier-Stokes equation for a viscous, incompressible fluid in vorticity form on the unit torus,
where $u \in C([0,T]; H^r_{\text{per}}((0,l)^2; \R^2))$ for any $r>0$ is the velocity field, $w = \nabla \times u$ is the vorticity, $w_0 \in L^2_{\text{per}}((0,l)^2;\R)$ is the initial vorticity,  $\nu \in \R_+$ is the viscosity coefficient, and $f \in L^2_{\text{per}}((0,l)^2;\R)$ is the forcing function. 
We want to learn the operator mapping the vorticity from the initial condition to the full solution
$\Gtrue: w_0 \mapsto w|_{[0,T]}$. 
\begin{align}
\label{eq:ns}
\begin{split}
\partial_t w(x,t) + u(x,t) \cdot \nabla w(x,t) &= \nu \Delta w(x,t) + f(x), \qquad x \in (0,l)^2, t \in (0,T]  \\
\nabla \cdot u(x,t) &= 0, \qquad \qquad \qquad \qquad \quad x \in (0,l)^2, t \in [0,T]  \\
w(x,0) &= w_0(x), \qquad \qquad \qquad \quad x \in (0,l)^2 
\end{split}
\end{align}
Specially, we consider two problem settings:
\begin{itemize}
    \item \textbf{Long temporal transient flow}: we study the build-up of the flow from the initial condition $u_0$  near-zero velocity to $u_T$ that reaches the ergodic state. We choose $t \in [0, 50]$, $l=1$, $Re=20$ as in \citet{li2020fourier}. The main challenge is to predict the long time interval.
    \item \textbf{Chaotic Kolmogorov flow}: In this case $u$ lies in the attractor where arbitrary starting time $t_0$. We choose $t \in [t_0, t_0 + 0.5] $ or $ [t_0, t_0 + 1]$, $ l=2\pi $, $Re=500$ similar to \citet{li2021markov}. The main challenge is to capture the small details that evolve chaotically.
    \item \textbf{Lid cavity flow}: In this case, we assume the no-slip boundary condition where $u(x,t) = (0, 0)$ at left, bottom, and right walls and $u(x,t) = (1, 0)$ on top, similar to \citet{bruneau20062d}.  We choose $t \in [5, 10]$, $ l=1 $, $Re=500$. The main challenge is to address the boundary using the velocity-pressure formulation.
\end{itemize}

\begin{figure}[t]
    \centering
    \includegraphics[width=14cm]{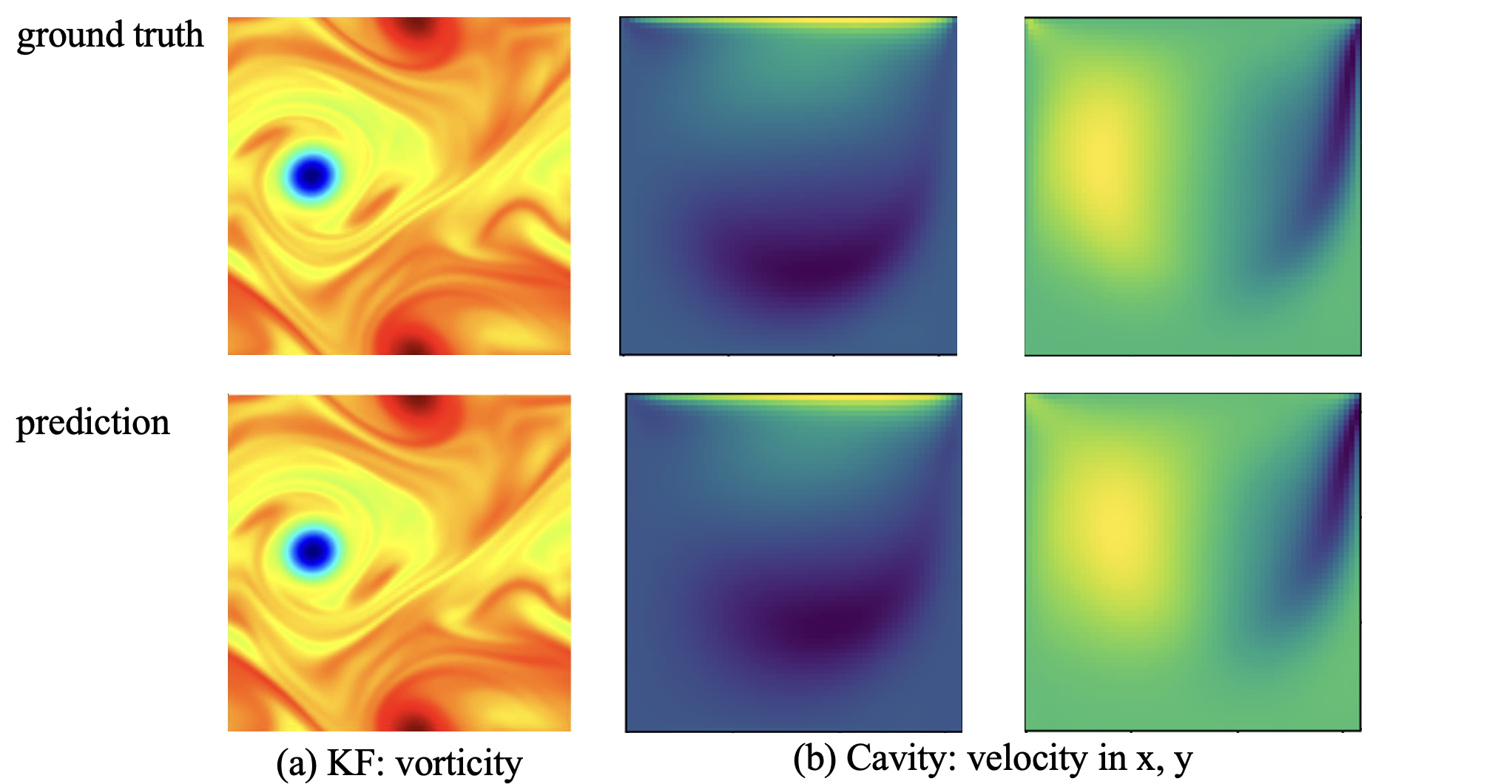}
    \caption{PINO on Kolmogorov flow (left) and Lid-cavity flow (right)}
    \label{fig:flow}
\end{figure}

\subsection{Operator learning with physics constraints}\label{sec:operator-learning}
We show that we can utilize the equation constraints to improve neural operator training. For this purpose, we train neural operators on fixed-resolution data in the presence of physics loss, $\Jpde$, and test the performance of the trained operators on high-resolution data. In particular, we test the performance of the trained model on data with the same resolution of the training data, 1x, 2x, and 4x, of the training data resolution~Table~\ref{tb:operator_resolution}. We observe that incorporating the $\Jpde$ in the training results in operators that, with high accuracy, generalize across data resolution. In this experiment, the training data for Burgers equation setting is in $32\times25$ (spatio-temporal), and the $\Jpde$ is imposed in $128\times100$ resolution. We use 800 low-resolution data and the same 800 PDE instances. The mean relative $L_2$ error and its standard deviation are reported over 200 test instances at resolution $32\times25$, $64\times50$, and $128\times100$. 

Accordingly, the training data for the Darcy equation setting is at the spatial resolution of $11\times11$ and the $\Jpde$ is imposed in $61\times 61$ resolution. We use 1000 low-resolution data and the same 1000 PDE instances. The mean and standard error are reported over 500 test instances at resolution $11\times11$, $61\times61$, and $211\times211$. Darcy flow is unresolved at the $11\times11$ resolution, training on such a coarse grid causes higher errors. However, adding higher resolution PDE loss helps the operator to resolve. 

The training data for Kolmogorov flow is in $64\times 64\times 33$ and the $\Jpde$ is imposed in $256\times 256 \times 65$ resolution for the time interval $[0, 0.125]$. We use 800 low-resolution data and 2200 PDE instances. The mean and std of the relative $L_2$ error are reported over 200 test instances, at resolution $64\times 64\times 33$, $128\times 128\times 33$, and $256\times 256\times 65$.

\begin{table}[htbp]
\centering
\begin{tabular}{|ll|c|c|c|c|}
\hline
 PDE & Training setting & \vtop{\hbox{\strut Error at low}\hbox{\strut data resolution}} & \vtop{\hbox{\strut Error at $2\times$}\hbox{\strut data resolution }} &\vtop{\hbox{\strut Error at $4\times$}\hbox{\strut data resolution}}\\ \hline
\hline
\multirow{2}{*}{Burgers}          & Data            & 0.32$\pm$0.01\%        & 3.32$\pm$0.02\% & 3.76$\pm$0.02\%         \\
& Data \& PDE loss        & 0.17$\pm$0.01\%        & 0.28$\pm$0.01\% & 0.38$\pm$0.01\%         \\
\hline
\multirow{2}{*}{Darcy}          & Data            & 5.41$\pm$0.12\%        & 9.01$\pm$0.07\% & 9.46$\pm$0.07\%         \\
& Data \& PDE loss        & 5.23$\pm$0.12\%        & 1.56$\pm$0.05\% & 1.58$\pm$0.06\%         \\
\hline
\multirow{2}{*}{Kolmogorov flow}         & Data              & 8.28\%$\pm$0.15\%        & 8.27\%$\pm$0.15\%&  8.30\%$\pm$0.15\%       \\
            & Data \& PDE loss          & 6.04\%$\pm$0.12\%       & 6.02\%$\pm$0.12\% &  6.01\%$\pm$0.12\%      \\ \hline
\end{tabular}\\
\caption{Operator-learning using fixed resolution data and PDE loss allows us to train operators with high accuracy on  higher resolution unseen data.}
\label{tb:operator_resolution}
\end{table}


\paragraph{Burgers equation and Darcy equation.} PINO can learn the solution operator without any data on simpler problems such as  Burgers and Darcy. Compared to other PDE-constrained operators, PINO is more expressive and thereby achieves better accuracy. On Burgers \eqref{eq:burgers}, PI-DeepONet achieves \textbf{1.38\%} \citep{wang2021learning}; PINO achieves \textbf{0.38\%}.  
Similarly, on Darcy flow \eqref{eq:darcy}, PINO outperforms FNO by utilizing physics constraints, as shown in Table \ref{tb:darcy-operator}. For these simpler equations, instance-wise fine-tuning may not be needed. The implementation detail and the search space of parameters are included in Appendix \ref{app:burgers} and \ref{app:darcy}.
\begin{table}[htbp]
\centering
\begin{tabular}{|l|c|}
\hline
Method         & Solution error  \\ \hline \hline
DeepONet with data \citep{lu2019deeponet}       &    $6.97\pm 0.09\%$                  \\ 
PINO with data & $1.22\pm 0.03\%$                        \\ \hline 
PINO w/o data  & $1.50\pm 0.03  \%$         \\ \hline
\end{tabular}
\caption{Operator learning on Darcy Flow equation. Incorporating physics constraints in operator learning improves the performance of neural operators.}
\label{tb:darcy-operator}
\end{table}

\begin{table}[htbp]
\centering
\begin{tabular}{|ll|l|}
\hline
\# data samples & \# PDE instances & Solution error  \\ \hline
\hline
0             & 2200            & 6.22\%$\pm$0.11\%            \\
\hline
800           & 2200               & 6.01\%$\pm$0.12\%           \\
\hline
2200            & 2200            & 5.04\%$\pm$0.11\%            \\ \hline
\end{tabular}\\
\caption{Physics-informed neural operator learning on Kolmogorov flow $Re=500$. PINO is effective and flexible in combining physics constraints and any amount of available data. 
The mean and standard error of the relative $L_2$ test error is reported over 200 instances and evaluated on resolution $256\times 256 \times 65$. }
\label{tb:operator-ansatz}
\end{table}

\paragraph{Chaotic Kolmogorov flow.}
We conduct an empirical study on how PINO can improve the generalization of neural operators by enforcing more physics. 
In the first experiment, we consider the Kolmogorov flow with $T=0.125$. We train PINO with 2200 initial conditions and different amounts of low-resolution data. As shown in Table~\ref{tb:operator-ansatz}, PINO achieves $6.22$\% error even without any data. We also observe that adding more low-resolution data to training makes the optimization easier and consistently improves the accuracy of the learned operator, showing that PINO is effective and flexible in combining physics constraints and any amount of available data. 

The second experiment considers the Kolmogorov flow with $T=0.5$. The training set consists of 4000 data points of the initial condition and corresponding solution. For operator learning, we sample high-resolution initial conditions from a Gaussian random field.
Table \ref{tb:operator-validation-4C} compare the generalization error of neural operators trained by different schemes and different amounts of simulated data. The result shows that training neural operator with additional PDE instances consistently improves the generalization error on all three resolutions we are evaluating. Note that the relative $L_2$ error in this setting is much higher than the first one because the time horizon is 4 times longer. Next, we show how to solve for specific instances by finetuning the learned operator.

\subsection{Solve equation using operator ansatz}\label{sec:solve-eqn}
We solve specific equation instances by fine-tuning the learned operator ansatz.
%

\paragraph{Long temporal transient flow.}
It is extremely challenging to propagate the information from the initial condition to future time steps over such a long interval $T=[0,50]$ just using the soft physics constraint. Neither the PINN nor PINO (from scratch) can handle this case (error $>50\%$), no matter solving the full interval at once or solving per smaller steps.
However, when the data is available for PINO, we can use the learned neural operator ansatz and the anchor loss $\Lanchor$.  The anchor loss is a hard constraint that makes the optimization much easier. Providing $N=4800$ training data, the PINO without instance-wise fine-tuning achieves \textbf{2.87\%} error, lower than FNO \textbf{3.04\%} and it retains a 400x speedup compared to the GPU-based pseudo-spectral solver \citep{he2007stability}, matching FNO.
Further doing test time optimization with the anchor loss and PDE loss, PINO reduces the error to \textbf{1.84\%}.

\paragraph{Chaotic Kolmogorov flow.}
Based on the solution operators learned in Section \ref{sec:operator-learning}, the second operator-learning setting, we continue to do instance-wise fine-tuning. We compare our method against other physics-informed learning methods including PINN \citep{raissi2019physics}, LAAF-PINN \citep{jagtap2020locally}, and SA-PINN \citep{mcclenny2020self}, as shown in Figure \ref{fig:pinn-baselines} and Table~\ref{tb:baseline_compare}. Overall, PINO outperforms PINN and its improved variants by \textbf{20x} smaller error and \textbf{25x} speedup. Using a learned operator model makes PINO converge faster. The implementation detail and the search space of parameters are included in Appendix \ref{app:kf}.
 
\begin{table}[htbp]
\centering
\begin{tabular}{|c|c|c|c|c|}
\hline
Method           & \# data samples &  \# PDE instances & Solution error $(w)$ & Time cost \\ \hline
\hline
PINNs & -  &      -      & 18.7\%                      &   4577s        \\ \hline
PINO   & 0 &  0  &\textbf{0.9\%}             & 608s      \\
PINO  & 0.4k &  0    & \textbf{0.9\%} & 536s\\ 
PINO  & 0.4k &  160k   & \textbf{0.9\%}                       & \textbf{473}s     \\ \hline
\end{tabular}
\caption{Instance-wise fine-tuning on Kolmogorov flow $Re=500$, $T = 0.5$. Using the learned operator as the initial condition helps fine-tuning converge faster. } 
\label{tb:baseline_compare}
\end{table}

\paragraph{Zero-shot super-resolution.}
The neural operator models are discretization-convergent, meaning they can take the training dataset of variant resolutions and be evaluated at higher resolution. As shown in Figure \ref{fig:spectrum50}, we train the FNO, PINO, and UNet model with $64\times 64 \times 32$ Kolmogorov Flows dataset and evaluate them at $256\times 256 \times 65$ resolution. Any frequencies higher than 64 are unseen during the training time. Conventional models such as UNet are not capable of direct super-resolution. For compassion, we equip it with tri-linear interpolation. For PINO, we also do test-time optimization. As shown in Figure \ref{fig:spectrum50}, the spectrums of the predictions are averaged over 50 instances. PINO with the test-time optimization achieves a very high accuracy rate, and its spectrum overlaps with the ground truth spectrum. However, Conventional models such as UNet+Interpolation have noising prediction with oscillating high frequencies. On the other hand, with the help of test-time optimization, PINO can extrapolate to unseen frequencies with high accuracy.

\paragraph{Transfer Reynolds numbers.}
The extrapolation of different parameters and conditions is one of the biggest challenges for ML-based methods. It poses a domain shift problem. In this experiment, we train the source operator model on one Reynolds number and then fine-tune the model to another Reynolds number, on the Kolmogorov flow with $T=1$. As shown in Table \ref{tb:transfer} by doing instance-wise fine-tuning, PINO can be easily transferred to different Reynolds numbers ranging from $100$ to $500$. 
This transferability shows PINO learned the dynamics shared across different Reynolds numbers.
Such property envisions broad applications including transferring the learned operator to different boundary conditions or geometries.

\paragraph{Lid cavity flow.}
We demonstrate an additional example using PINO to solve for lid-cavity flow on $T=[5,10]$ with $Re=500$. In this case, we do not have the operator-learning phase and directly solve the equation (instance-wise fine-tuning). We use PINO with the velocity-pressure formulation and resolution $65\times 65\times 50$ plus the Fourier numerical gradient. It takes $2$ minutes to achieve a relative error of $14.52\%$.  Figure \ref{fig:flow} shows the ground truth and prediction of the velocity field at $t=10$ where the PINO accurately predicts the ground truth. The experiment shows that PINO can address non-periodic boundary conditions and multiple output fields.

\paragraph{Convergence of accuracy with respect to resolution.} 
We study the convergence rate of PINO in the instance-wise optimization setting, where we minimize the PDE loss under different resolutions without any data.  
For PINO, using a higher resolution is more effective compared to running gradient descent for longer iterations.
We test PINO on the Kolmogorov flow with Re =
500 and T = 0.125. We use the Fourier method in the spatial dimension and the finite difference method in the temporal dimension. 
As shown in Table \ref{table:convergence}, PINO shares the same convergence rate of its differentiation methods with no
obvious limitation from optimization. It has a first-order convergence rate in time when $dx$ is fine enough and an exponential convergence rate when $dt$ is fine enough. It implies the PDE constraint can achieve high accuracy given a reasonable computational cost, and the virtual instances are almost as good as the data instances generated by the solver. Since the PDE loss can be computed on an unlimited amount of virtual instances in the operator learning setting, it is possible to reduce the generalization error going to zero by sampling virtual instances.

\begin{table}[h!]
\centering
\begin{tabular}{|l|c|c|c|c|c|}
\hline
\diagbox{dx\hspace{.5cm}}{\hspace{2em}\raisebox{-.3cm}{dt}}&
$2^{-6}$& $2^{-7}$ & $2^{-8}$ & $2^{-9}$ & $2^{-10}$\\
\hline
$2^{-4}$ & 0.4081 & 0.3150  & 0.3149 &  0.3179 & 0.3196 \\
\hline
$2^{-5}$ & 0.1819 & 0.1817 & 0.1780 & 0.1773 & 0.1757\\
\hline
$2^{-6}$ & 0.0730 & 0.0436 & 0.0398 &  0.0386 & 0.0382\\
\hline
$2^{-7}$ & 0.0582 & 0.0234 & 0.0122 & 0.0066 & 0.0034\\
\hline  

\end{tabular}
\caption{relative L2 error of PINO (Finite-difference in time) on Kolmogorov flow with $Re=500$ and $T=0.125$. PINO inherits the convergence rate of its differentiation method with no limitation of optimization}
\label{table:convergence}
\end{table}

\begin{figure}
\centering
    \begin{subfigure}[t]{0.25\columnwidth}
    \includegraphics[width=1.0\textwidth]{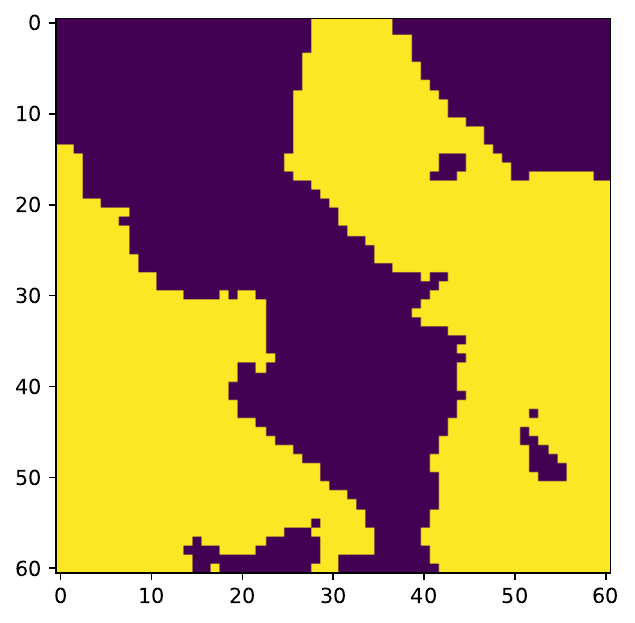}
    \caption{Ground truth input $a$}
    \label{fig:true-input}
    \end{subfigure}
    \begin{subfigure}[t]{0.25\columnwidth}
    \includegraphics[width=1.0\textwidth]{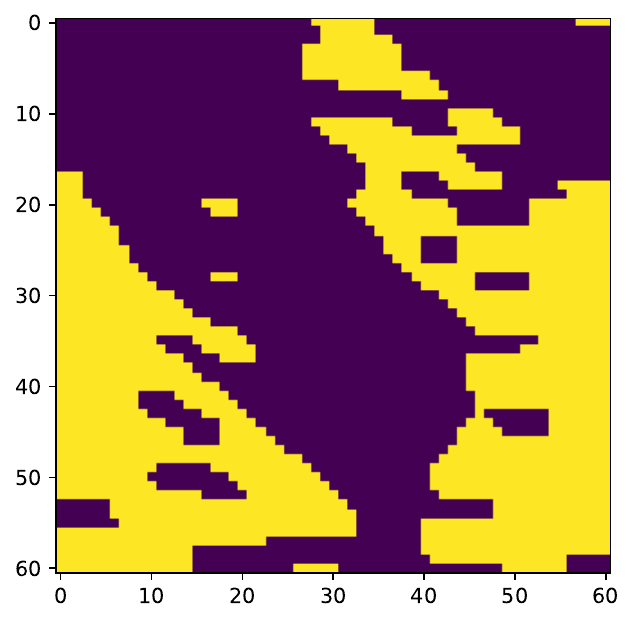}
    \caption{Inversion using only data constraint}
    \label{fig:inverse_data-input}
    \end{subfigure}
    \begin{subfigure}[t]{0.25\columnwidth}
    \includegraphics[width=1.0\textwidth]{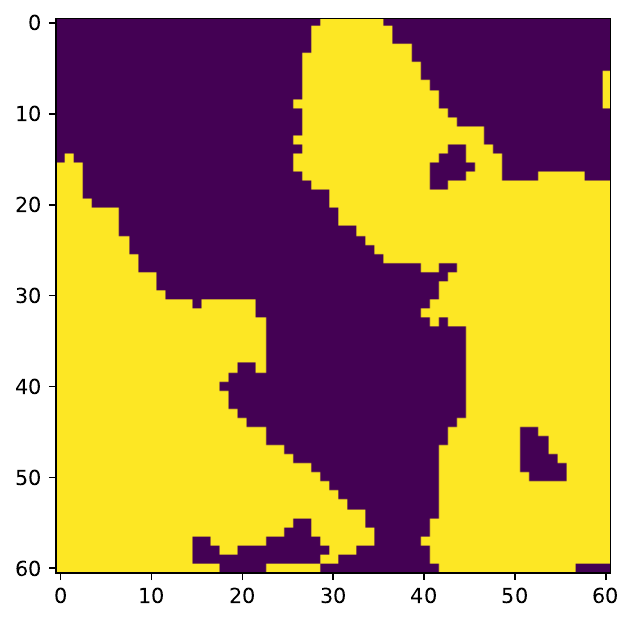}
    \caption{Inversion using data and PDE constraints}
    \label{fig:inverse_data_PDE-input}
    \end{subfigure}
    \begin{subfigure}[t]{0.25\columnwidth}
    
    \includegraphics[width=1.0\textwidth]{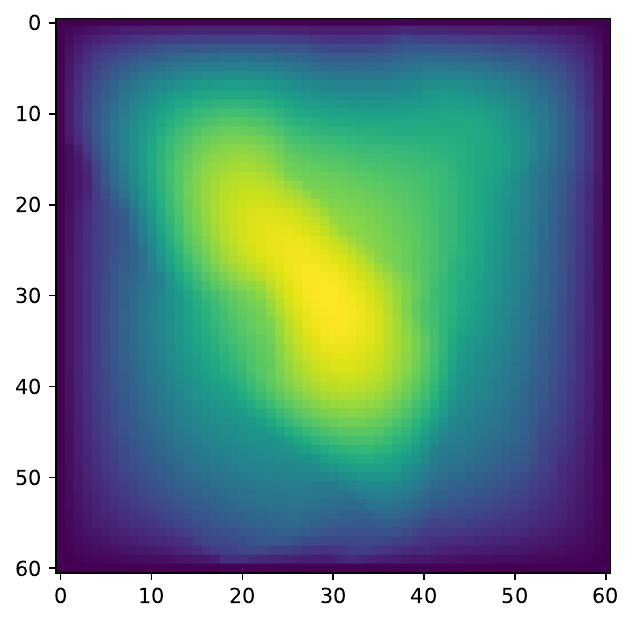}
    \caption{Observed output function}
    \label{fig:true-output}
    \end{subfigure}
    \begin{subfigure}[t]{0.25\columnwidth}
    \includegraphics[width=1.0\textwidth]{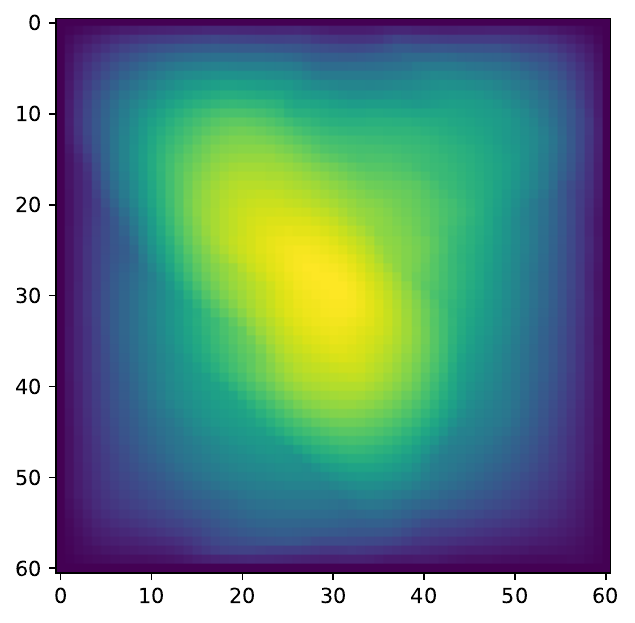}
    \caption{Output function of inversion using only data constraint}
    \label{fig:inverse_data-output}
    \end{subfigure}
    \begin{subfigure}[t]{0.25\columnwidth}
    \includegraphics[width=1.0\textwidth]{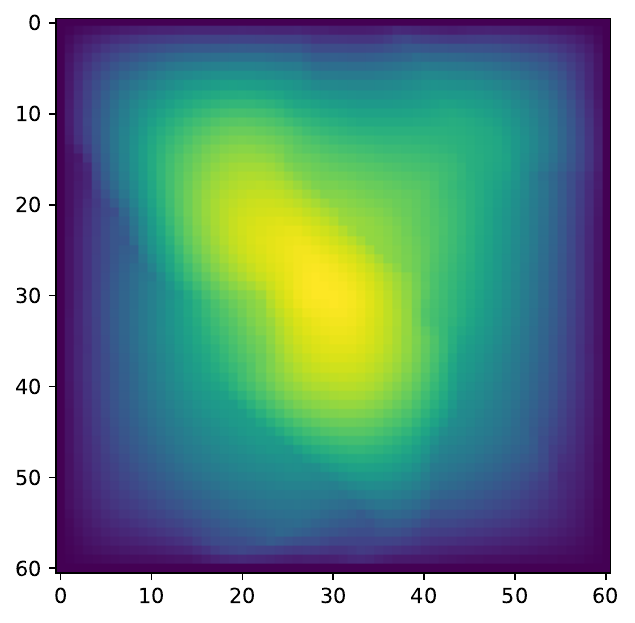}
    \caption{Output function of inversion using data and PDE constraints}
    \label{fig:inverse_data_PDE-output}
    \end{subfigure}
    \caption{In the above figures, (\ref{fig:true-input}) represents the ground truth input function $a^\dagger$, and (\ref{fig:true-output}) demonstrates the corresponding solution $u^\dagger$, i.e., the output function. Given the output $u^\dagger$, we aim to recover what $a$ could have generated the output function $u^\dagger$. Using only data constraint, (\ref{fig:inverse_data-input}) shows that our method can find an $a$ that results in an output function very close to the ground truth $u^\dagger$~(\ref{fig:inverse_data-output}). However, the recovered $a$ is far from satisfying the PDE equation. Using both data and PDE constraints, (\ref{fig:inverse_data_PDE-input}) shows that our physics-informed method can find an $a$ that not only results in an output function very close to the ground truth $u^\dagger$(\ref{fig:inverse_data_PDE-output}), but also the recovered $a$ satisfies the PDE constraint and is close to the underlying $a^\dagger$. }
    \label{fig:inverseproblem_pde}
\end{figure}

\begin{figure}[t]
    \centering
    \includegraphics[width=12cm]{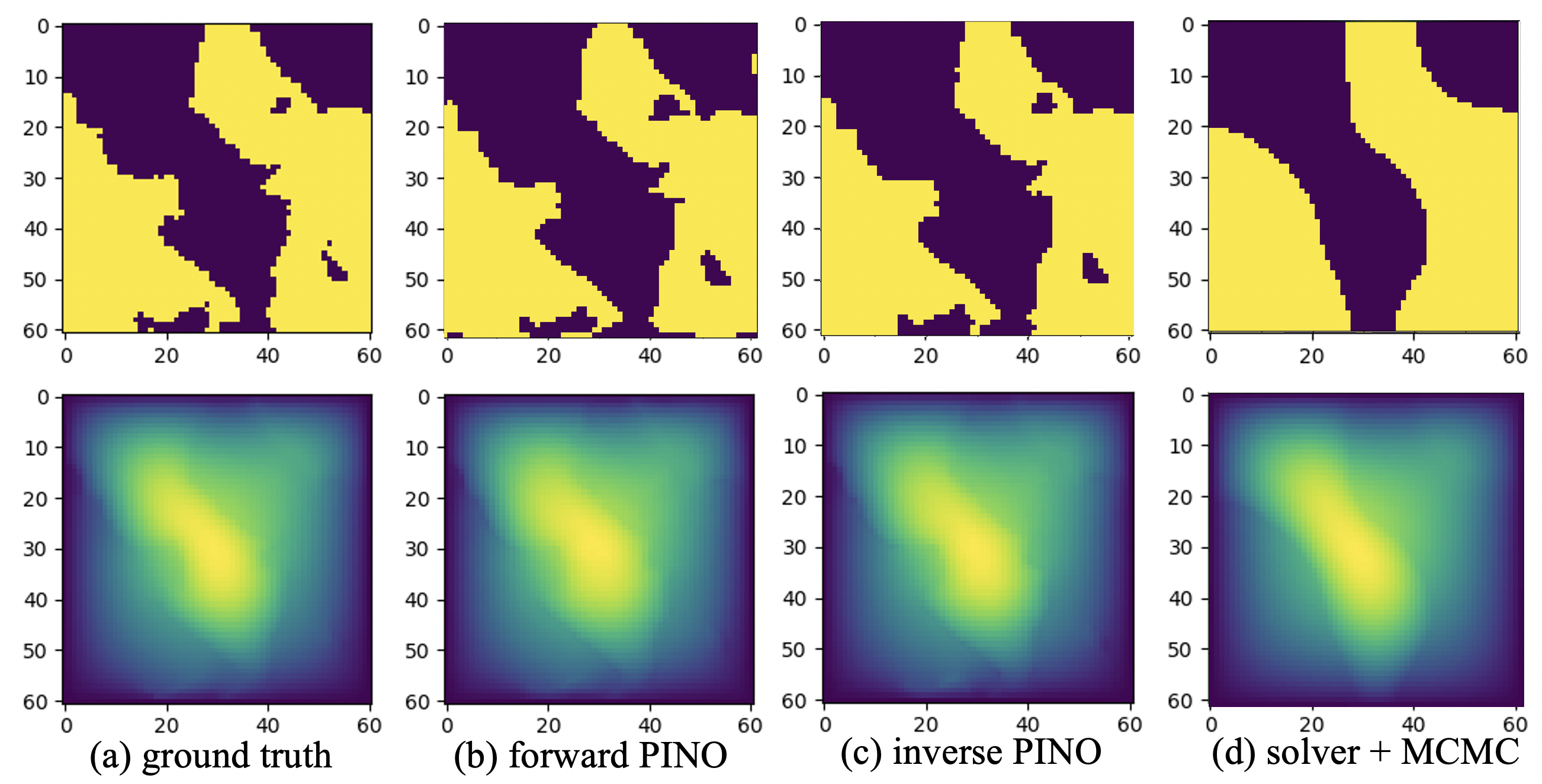}
    \caption{Darcy inverse problem: comparing PINO forward, inverse models with numerical solver with MCMC. }
    \label{fig:IP}
\end{figure}

\subsection{Inverse problem}\label{sec:exp-inverse-problem}
One of the major advantages of the physics-informed method is to solve the inverse problem. 
In the experiment, we investigate PINO on the inverse problem of the Darcy equation to recover the coefficient function $a^{\dagger}$ from the given solution function $u^{\dagger}$. 
We assume a dataset $\{a_j,u_j\}$ is available to learn the operator.
The coefficient function $a$ is a piecewise constant (representing two types of media), so the inverse problem can be viewed as a classification problem. We define $R(a)$ as the total variance. 

The PDE loss strongly improves the prediction of the inverse problem.
The plain neural operator model, while accurate in the forward problem, is vulnerable under perturbation and shift of the input $a$, which is a common behavior of deep-learning models. 
This domain-shift problem is critical for optimization-based inverse problems.
During the optimization, $a$ is likely to go out of the training distribution, which makes the neural operator model inaccurate. As shown in Figure \ref{fig:inverseproblem_pde} (b), the prediction of $a$ is less accurate, while the model believes its output \ref{fig:inverseproblem_pde} (e) is the same as the target.
This issue is mitigated by adding the PDE constraints, which restrict the prediction $a$ to the physically-valid manifold where $\mathcal{P}(a,u)=0$. As shown in Figure \ref{fig:inverseproblem_pde} (c), 
the initial condition recovered with PDE loss is very close to the ground truth. 

Comparing the PINO forward model with the inverse model, 
the inverse model  $\F_\theta: u \mapsto a$ \eqref{eq:backward} has the best performance. As Shown in Figure \ref{fig:IP}, the inverse model has \textbf{2.29\%} relative l2 error on the output $u$ and \textbf{97.10\%} classification accuracy on the input $a$; the forward model has 6.43\% error on the output and 95.38\% accuracy on the input. Both models converge with 200 iterations. 
The major advantage of the PINO inverse model compared to the PINO forward model is that it uses a neural operator $\F_\theta(u^{\dagger})$ as the ansatz for the coefficient function, which is used as regularization  $\Lanchor$. Similar to the forward problem, the operator ansatz has an easier optimization landscape while being expressive.

As a reference, we compare the PINO inverse frameworks with PINN  and the conventional solvers using the accelerated MCMC method with 500,000 steps \citep{cotter2013mcmc}.
The posterior mean of the MCMC has a 4.52\% error and 90.30\% respectively ( Notice the Bayesian method outputs the posterior distribution, which is beyond obtaining a maximum a posteriori estimation). 
Meanwhile, PINO methods are \textbf{3000x} faster compared to MCMC 
PINN does not converge in this case.
Please refer to \ref{sec:invsers_problem_ablation} for further study on the importance of imposing physics constraints in approaching inverse problems in PDE.

Besides the speedup with respect to the online cost, the offline training of PINO only takes around 1 hour on a single GPU on the Darcy problem. Once trained, the model can be used without any further training cost. 
As a comparison, it takes considerably longer to deploy a finite element solver and an MCMC solver compared to a machine learning model. Generally speaking, numerical solvers usually have a more complicated codebase and it is non-trivial to specify boundary conditions, time schemes, and meshes. In the end, it can be easier to prepare a machine learning model than a standard numerical solver. Flexibility and accessibility are some of the major advantages of these machine learning methods.

%% file: sections/6-discussion.tex
\section{Conclusion and future work}

In this work, we develop the physics-informed neural operator (PINO) that bridges the gap between physics-informed optimization and data-driven neural operator learning. We introduce operator-learning and instance-wise fine-tuning schemes for PINO to utilize both the data and physics constraints. In the operator learning phase, PINO learns an operator ansatz over multiple instances of a parametric PDE family. The instance-wise fine-tuning scheme allows us to take advantage of the learned neural operator ansatz and solve for the solution function on the querying instance faster and more accurately.

While PINO shows many promising applications, it also shares some limitations as in the previous work. For example, since PINO is currently implemented with the FNO backbone with the Fast-Fourier transform, it is hard to extend to higher dimensional problems. Besides, as shown in Figure \ref{fig:kf}, finetuning PINO using gradient descent methods does not converge as fast as using a finer grid as in Table \ref{table:convergence}. Further optimization techniques are to be developed.

There are many exciting future directions. Most of the techniques and analyses of PINN can be transferred to PINO.  It is also interesting to ask how to overcome the hard trade-off of accuracy and complexity, and how the PINO model transfers across different geometries. Furthermore, it is promising to develop a software library of pre-trained models. PINO's excellent extrapolation property allows it to be applied on a broad set of conditions, as shown in Transfer Reynold's number experiments.

%% file: sections/Appendix.tex
\newpage

\section{Implementation details}\label{sec:appendix}

In this section, we list the detailed experiment setups and parameter searching for each experiment in Section \ref{sec:exp}. Without specification, we use Fourier neural operator backbone with $width = 64$, $mode = 8$ or $12$, $L=4$, and GeLU activations. The numerical experiments are performed on Nvidia V100 GPUs and A100 GPUs. 

\subsection{Burgers Equation}\label{app:burgers}
We use the $1000$ initial conditions \(u_0 \sim \mu\) where \(\mu = \mathcal{N}(0,625(-\Delta + 25I)^{-2})\)   to train the solution operator on PINO with $width = 64$, $mode = 15$, and GeLU activation. We use the numerical method to take the gradient. We use Adam optimizer with the learning rate $0.001$ that decays by half every $100$ epochs. $500$ epochs in total. The total training time is about $20$ minutes on a single Nvidia 3090 GPU. PINO achieves \textbf{0.38\%} relative l2 error averaged over 200 testing instances.  PINN-DeepONet achieves \textbf{1.38\%} which is taken from \citet{wang2021learning} which uses the same problem setting.

\subsection{Darcy Flow}\label{app:darcy}
We use the $1000$ coefficient conditions $a$ to train the solution operator where \(a \sim \mu \) where \(\mu = \psi_{\#} \mathcal{N}(0,(-\Delta + 9I)^{-2})\), $\psi(a(x)) = 12$ if $ a(x)\geq 0$; $\psi(a(x)) = 3 $ if $ a(x)<0$. The zero boundary condition is enforced by multiplying a mollifier $m(x) = \sin(\pi x)\sin(\pi y)$ for all methods.
The parameter of PINO on Darcy Flow is the same as in the Burgers equation above.
It takes around 1 hour to train the PINO model on a single Nvidia V100 GPU.
Regarding the implementation detail of the baselines: as for FNO, we use the same hyperparameters as its paper did \citep{li2020neural} and set $width=64, mode=20, depth=4$; DeepONet \citep{lu2019deeponet} did not study Darcy flow so we grid search the hyperparameters of DeepONet: depth from 2 to 12, width from 50 to 100. The best result of DeepONet is achieved by depth 8, and width 50. The results are shown in Table \ref{tb:darcy-operator}. All the models are trained on resolution $61\times 61$ and evaluated on resolution $61\times 61$.


\subsection{Long temporal transient flow.}\label{app:long}
We study the build-up of the flow from the initial condition $u_0$  near-zero velocity to $u_T$ that reaches the ergodic state. We choose $T = 50, l=1$ as in \citet{li2020fourier}. We choose the weight parameters of error $\alpha=\beta=5$.
The initial condition $ w_0(x)$ is generated according to 
\(w_0 \sim \mu\) where \(\mu = \mathcal{N}(0,7^{3/2}(-\Delta + 49I)^{-2.5})\) with periodic boundary conditions. The forcing is kept fixed $f(x) = 0.1(\sin(2\pi(x_1+x_2)) + \cos(2\pi(x_1 + x_2)))$. We compare FNO, PINO (no instance-wise fine-tuning), and PINO (with instance-wise fine-tuning). They get $3.04\%$, $2.87\%$, and $1.84\%$ relative l2 error on the last time step $u(50)$ over 5 testing instance.



\subsection{Chaotic Kolmogorov flow.}\label{app:kf}
For this experiment,  $u$ lies in the attractor. We choose $T =0.125, 0.5 $ or $ 1$, and $ l=1 $, similar to \citet{li2021markov}. 
For $T=0.5s$, the training set consists of 4000 initial condition functions and corresponding solution functions with a spatial resolution of $64\times 64$ and a temporal resolution of $65$. Extra initial conditions are generated from Gaussian random field $\mathcal{N}\left(0,7^{3/2}(-\Delta+49I)^{-5/2}\right)$.
We estimate the generalization error of the operator on a test set that contains 300 instances of Kolmogorov flow and reports the average relative $L_2$ error. 
Each neural operator is trained with 4k data points plus a number of extra sampled initial conditions. The Reynolds number in this problem is 500. The reported generalization error is averaged over 300 instances. 
It takes up to 2 days to train the model on a single Nvidia V100 GPU.
For the experiments in Table~\ref{tb:operator_resolution}, $T=0.125$ and the training set has data with spatial resolution $64\times 64$ and temporal resolution $33$. The test set has 200 instances with spatial resolution $256\times 256$ and temporal resolution $65$. For the experiments in Table~\ref{tb:low_res_kf}, $T=0.125$ and the training set contains 800 instances in even lower resolution $32\times 32 \times 17$. The test set has 200 instances with spatial resolution $256\times 256$ and temporal resolution $65$.

\paragraph{Comparison study.}\label{pinn}
The baseline method PINN, LAAF-PINN, and SA-PINN are implemented using library DeepXDE \citep{lu2021deepxde} with TensorFlow as the backend. 
We use the two-step optimization strategy (Adam \citep{kingma2014adam} and L-BFGS) following the same practice as NSFNet \citep{jin2021nsfnets}, which applies PINNs to solving Navier Stokes equations. We grid search the hyperparameters: network depth from 4 to 6, width from 50 to 100, learning rate from 0.01 to 0.0001, and the weight of boundary loss from 10 to 100 for all experiments of PINNs. 
Comparison between PINO and PINNs on instance-wise fine-tuning. The results are averaged over 20 instances of the Navier-Stokes equation with Reynolds number 500. The best result is obtained by PINO using learned operator ansatz and virtual sampling. The neural operator ansatz used here is trained on 400 data points.
The authors acknowledge that there could exist more sophisticated variants of PINN that perform better in our test cases. 

\begin{table}[ht]
\centering
\begin{tabular}{|c|c|c|}
\hline
Test resolution & FNO                         & PINO \\ \hline \hline
64x64x33        & 9.73$\pm$ 0.15\% &    6.30$\pm$0.11\%  \\ \hline
128x128x33      & 9.74$\pm$ 0.16\% &   6.28$\pm$0.11\%  \\ \hline
256x256x65      & 9.84$\pm$ 0.16\% &  6.22$\pm$0.11\% \\ \hline
\end{tabular}
\caption{Comparison between data only (FNO) and data + PDE (PINO) on higher resolutions while trained on much lower resolution $32\times 32\times 17$. }
\label{tb:low_res_kf}
\end{table}


\begin{table}[htbp]
\centering
\begin{tabular}{|l|l|l|l|l|}
\hline
\# data samples       & \# additional PDE instances       & Resolution & Solution error & Equation error \\ \hline \hline
\multirow{3}{*}{400} & \multirow{3}{*}{0}    & $128\times 128\times 65$ & 33.32\%              & 1.8779                  \\
                     &                       & $64\times 64\times 65$   & 33.31\%              & 1.8830                  \\
                     &                       & $32\times 32\times 33$   & 30.61\%              & 1.8421                  \\ \hline
\multirow{3}{*}{400} & \multirow{3}{*}{40k}  & $128 \times 128 \times 65$ & 31.74\%              & 1.8179                  \\
                     &                       & $64\times 64\times 65$   & 31.72\%              & 1.8227                  \\
                     &                       & $32\times 32\times 33$   & 29.60\%              & 1.8296                  \\ \hline
                     
\multirow{3}{*}{400} & \multirow{3}{*}{160k} & $128\times 128\times 65$ & {31.32\%}              & 1.7840                  \\
                     &                       & $64\times 64\times 65$   & {31.29\%}              & 1.7864                \\
                     &                       & $32\times 32\times 33$   & {29.28\%}              & 1.8524                  \\ \hline\hline
                     
\multirow{3}{*}{4k} & \multirow{3}{*}{0}             & $128\times 128 \times 65$ & 25.15\%               & 1.8223                  \\
                    &           & $64\times 64 \times 65$   & 25.16\%              & 1.8257                  \\
                    &           & $32\times 32\times 33$   & 21.41\%               & 1.8468                  \\ \hline
\multirow{3}{*}{4k} & \multirow{3}{*}{100k}          & $128\times 128 \times 65$ & 24.15\%               & 1.6112                  \\
                     &          & $64\times 64 \times 65$   & 24.11\%              & 1.6159                  \\
                  &             & $32\times 32\times 33$    & 20.85\%              & 1.8251                  \\ \hline
\multirow{3}{*}{4k} & \multirow{3}{*}{400k}          & $128\times 128 \times 65$ & 24.22\%               & 1.4596                  \\
                  &             & $64\times 64 \times 65$   & 23.95\%              & 1.4656                  \\
                &           & $32\times 32\times 33$    & 20.10\%               & 1.9146                  \\ \hline\hline
\multirow{3}{*}{0} & \multirow{3}{*}{100k} & $128\times 128 \times 65$ & 74.36\%              & 0.3741                  \\
                  &             & $64\times 64 \times 65$   & 74.38\%             & 0.3899                 \\
                  &             & $32\times 32\times 33$    & 74.14\%             & 0.5226                  \\ \hline
\end{tabular}
\caption{Each neural operator is trained with 400 or 4000 data points additionally sampled free initial conditions. The Reynolds number is 500. The reported generalization error is averaged over 300 instances. Training on additional initial conditions boosts the generalization ability of the operator.}
\label{tb:operator-validation-4C}
\end{table}


\subsection{Transfer learning across Reynolds numbers}
\begin{figure}
    \centering
    \includegraphics[width=0.4\textwidth]{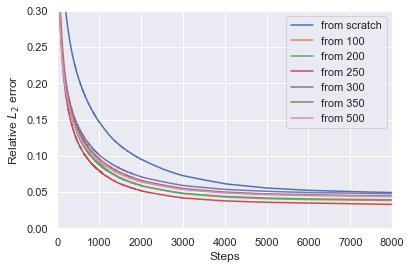}
    \caption{Plot of relative $L_2$ error versus update step for the Kolmogorov flow with Reynolds number $500$, $T=1$. The test error is averaged over 40 instances. We observe that all the operator ansatzs trained over PDE instances with different Reynolds numbers can boost the instance-wise fine-tuning accuracy and convergence speed compared to training from scratch.}
    \label{fig:test_curve_transfer}
\end{figure}

We study the instance-wise fine-tuning with different Reynolds numbers on the $T=1$ Kolmogorov flow.  For the higher Reynolds number problem $Re=500, 400$, fine-tuning the source operator shows better convergence accuracy than learning from scratch. In all cases, the fine-tuning of the source operator shows better convergence speed as demonstrated in Figure \ref{fig:test_curve_transfer}. The results are shown in Table \ref{tb:transfer} where the error is averaged over 40 instances. Each row is a testing case, and each column is a source operator. 

\begin{table}[htbp]
\centering
\begin{tabular}{|c|l|lllllll|}
\hline
Testing Re & From scratch   & 100            &  200            & 250            &  300            & 350            & 400            & 500            \\ \hline \hline
500             & 0.0493  & 0.0383  & 0.0393  & 0.0315  & 0.0477  & 0.0446  & 0.0434  & 0.0436  \\
400             & 0.0296  & 0.0243  & 0.0245  & 0.0244  & 0.0300  & 0.0271  & 0.0273  & 0.0240  \\
350             & 0.0192  & 0.0210  & 0.0211  & 0.0213  & 0.0233  & 0.0222  & 0.0222 & 0.0212  \\
300             & 0.0168  & 0.0161  & 0.0164  & 0.0151  & 0.0177  & 0.0173  & 0.0170  & 0.0160  \\
250             & 0.0151  & 0.0150  & 0.0153  & 0.0151  & 0.016  & 0.0156  & 0.0160  & 0.0151  \\
200             & 0.00921 & 0.00913 & 0.00921 & 0.00915 & 0.00985 & 0.00945 & 0.00923 & 0.00892 \\
100             & 0.00234 & 0.00235 & 0.00236 & 0.00235 & 0.00239 & 0.00239 & 0.00237 & 0.00237 \\ \hline
\end{tabular}
\caption{Reynolds number transfer learning. Each row is a test set of PDEs with corresponding Reynolds number. Each column represents the operator ansatz we use as the starting point of instance-wise fine-tuning. For example, column header ``100'' means the operator ansatz is trained over a set of PDEs with Reynolds number 100. 
The relative $L_2$ errors is averaged over 40 instances of the corresponding test set. }
\label{tb:transfer}
\end{table}

\section{Additional experiments}


\begin{figure}[t]
     \centering
     \includegraphics[width=14cm]{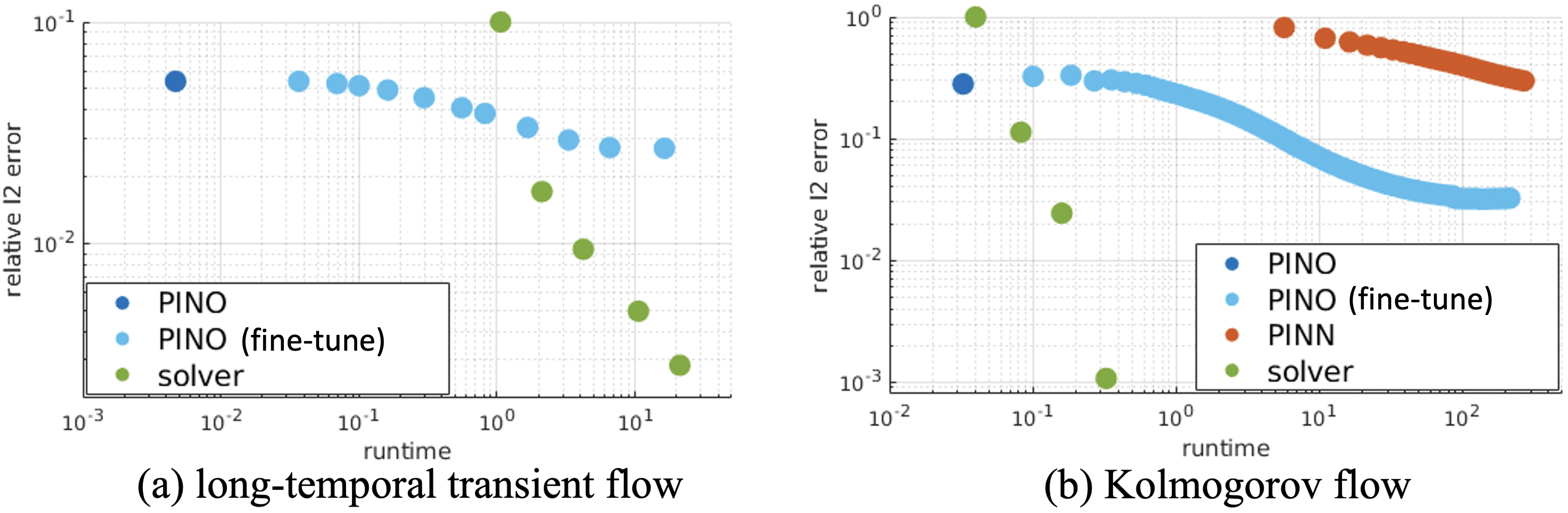}\\
     (a) The long-temporal transient flow with $Re\sim 20, T=50$. PINO outputs the full trajectory in one step, which leads to a 400x speedup compared to the GPU solver. PINN cannot converge to a reasonable error rate due to the long time window. (b) The chaotic Kolmogorov flow with $Re =\sim 500, T=0.5$. PINO converges faster compared to PINN, but their convergence rates with gradient descent are less effective compared to using higher resolutions in the GPU solver.
     \caption{The accuracy-complexity trade-off on PINO, PINN, and the GPU-based pseudo-spectral solver.}
     \label{fig:kf}
\end{figure}

\subsection{Additional baselines}

We add a comparison experiment against the Locally adaptive activation functions for PINN (LAAF-PINN) \citep{jagtap2020locally} and Self-Adaptive PINN (SA-PINN) \citep{mcclenny2020self}. For the Kolmogorov flow problem, we set Re=500, T=[0, 0.5].
We search among the following hyperparameters combinations:
LAAF-PINN: n: {10, 100}, learning rate: {0.1, 0.01, 0.001}, depth {4, 6}.
SA-PINNs: learning rate {0.001, 0.005, 0.01, 0.05}, 
network width {50, 100, 200}, depth {4, 6, 8}.

As shown in Figure \ref{fig:pinn-baselines}, both LAAF-PINN and SA-PINN converge much faster than the original PINN method, but there is still a big gap with PINO. LAAF-PINN adds learnable parameters before the activation function; SA-PINN adds the weight parameter for each collocation point. These techniques help to alleviate the PINNs` optimization problem significantly. However, they didn't alter the optimization landscape effectively in the authors' opinion. On the other hand, by using the operator ansatz, PINO optimizes function-wise where the optimization is fundamentally different.

Note that the contribution of PINO is orthogonal to the above methods. One can apply the adaptive activation functions or self-adaptive loss in the PINO framework too. All these techniques of PINNs can be straightforwardly transferred to PINO. We believe it would be interesting future directions to study how all these methods work with each other in different problems.

\subsection{Lid Cavity flow.}\label{app:cavity}

We demonstrate an addition example using PINO to solve for lid-cavity flow on $T=[5,10]$ with $Re=500$. In this case, we do not have the operator-learning phase and directly solve the equation (instance-wise fine-tuning). We use PINO with the velocity-pressure formulation and resolution $65\times 65\times 50$ plus the Fourier numerical gradient. It takes $2$ minutes to achieve a relative error of $14.52\%$.  Figure \ref{fig:flow} shows the ground truth and prediction of the velocity field at $t=10$ where the PINO accurately predicts the ground truth. 

We assume a no-slip boundary where $u(x,t) = (0, 0)$ at left, bottom, and right walls and $u(x,t) = (1, 0)$ on top, similar to \citet{bruneau20062d}.  We choose $t \in [5, 10]$, $ l=1 $, $Re=500$.  We use the velocity-pressure formulation as in \citet{jin2021nsfnets} where the neural operator output the velocity field in $x$, $y$, and the pressure field. We set $width = 32$, $mode = 20$ with a learning rate $0.0005$ which decreases by half every $5000$ iterations, $5000$ iterations in total. We use the Fourier method with Fourier continuation to compute the numerical gradient and minimize the residual error on the velocity, the divergence-free condition, as well as the initial condition and boundary condition. The weight parameters ($\alpha, \beta$) between different error terms are all chosen as 1. Figure \ref{fig:flow} shows the ground truth and prediction of the velocity field at $t=10$ where the PINO accurately predicts the ground truth.

\section{Fourier continuation}
\label{app:fc}
\begin{figure}
    \centering
    \includegraphics[width=1\textwidth]{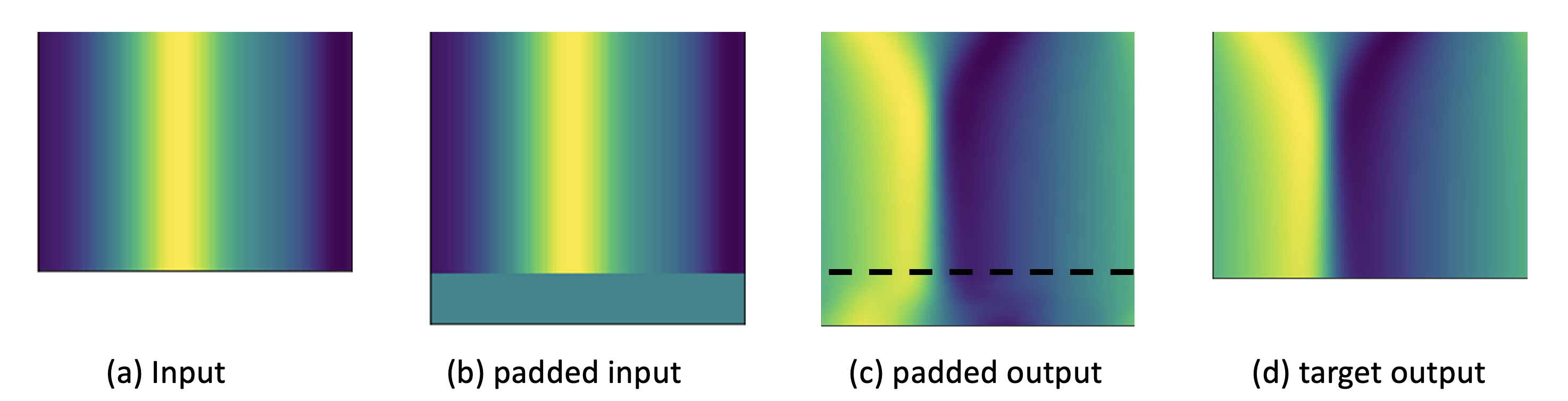}
    \caption{Fourier Continuation by padding zeros. The x-axis is the spatial dimension; the y-axis is the temporal dimension. FNO extends the output smoothly on the padded domain.}
    \label{fig:fc}
\end{figure}

The Fourier neural operator can be applied to arbitrary geometry via Fourier continuations. Given any compact manifold $\mathcal{M}$, we can always embed it into a periodic cube (torus), 
\[i: \mathcal{M} \to \mathcal{T}^n\]
where we can do the regular FFT. Conventionally, people would define the embedding $i$ as a continuous extension by fitting polynomials \citep{bruno2007accurate}. However, in Fourier neural operator, it can be simply done by padding zeros in the input. The loss is computed at the original space during training. The Fourier neural operator will automatically generate a smooth extension to do a padded domain in the output, as shown in Figure \ref{fig:fc}.

This technique is first used in the original Fourier neural operator paper \citep{li2020fourier} to deal with the time dimension in the Navier-Stokes equation. Similarly, \citet{lu2021comprehensive} apply FNO with extension and interpolation on diverse geometries on the Darcy equation. In the  work, we use Fourier continuation widely for non-periodic boundary conditions (Darcy, time dimension). We also added an example of lid-cavity to demonstrate that PINO can work with non-periodic boundary conditions.

Furthermore, this Fourier continuation technique helps to take the derivatives of the Fourier neural operator. Since the output of FNO is always on a periodic domain, the numerical Fourier gradient is usually efficient and accurate, except if there is shock (in this case, we will use the exact gradient method). 

\section{PDE constraints in inverse problems}\label{sec:invsers_problem_ablation}
In subsection~\ref{subsec:inverse_problem} we presented the study of inverse problems where we propose a new approach to tackle inverse problems. We propose to incorporate both data and PDE constraints for the inverse problem. The data constraint makes sure the recovered input function, when fed to the neural operator, results in an output that matches the observed data $u^\dagger$. The PDE constraint, imposed using $\Lpde$, is a crucial physics-based constraint that is the key to having an accurate approach for inverse problems. This constraint reinforces that the pairs of recovered input and their corresponding output are physical. In other words, they satisfy the underlying PDE. When the physics-based constraint is imposed, the search space of $(a,u)$ is confined to the solutions manifold of the PDE, i.e., pairs satisfying the underlying PDE.

\section{Discretization convergence and representation equivalence}


In this paper, we follow the definition of resolution and discretization convergence (previously named discretization invariance) provided in the definition of neural operators established in \citep{kovachki2021neural}. This definition states resolution convergence in the limit of mesh refinement. As the mesh refinement goes to infinity, a discretization convergent operator converges to its limit in the infinite-dimensional function space. It may get a higher error at a coarse discretization, but as the training resolution increases, the model should converge to an accurate model.

Another class of neural operator, named representation equivalent neural operator, is proposed later in \citep{bartolucci2023neural}. Representation equivalent neural operators are defined as these neural operators with no aliasing error. Such neural operators are invariant under any discretization.
FNOs do not satisfy representation equivalence because it has a pointwise non-linearity in every layer applied in physical space which can re-introduce Fourier modes of size greater than the current grid so, under the assumption that all modes are active for the input discretization, this will necessarily introduce aliasing error. This remains true for any architecture which uses pointwise non-linearities in physical space, encompassing most of the deep learning. 

Hence, resolution convergent \citep{kovachki2021neural} and representation equivalence \citep{bartolucci2023neural} are different goals to achieve. The former holds for FNO and PINO, which have the expressive power to approximate the underlying operator as the resolution goes to infinity. This allows them to make predictions at any resolution and hence, they can extrapolate to higher resolutions than their training data. On the other hand, representation equivalent models such as SNO \citep{fanaskov2022spectral} and PCA-Net \citep{Kovachki} are limited to fixed-dimensional representation, and they cannot generate higher frequencies beyond what they are trained on. 

As shown in the figure below, PINO trained on 64x64 resolution data can extrapolate to unseen higher frequencies. In contrast, representation-equivalent models cannot generate new frequencies since their representation space is fixed. Therefore they introduce an irreducible approximation error based on the size of the predefined representation space. Since the goal of operator learning is to find the underlying solution operator in the continuum, we believe discretization convergent is a more useful property.